Original papers

# Kriging prior regression: A case for kriging-based spatial features with TabPFN in soil mapping

Jonas Schmidinger [a,b,*], Viacheslav Barkov [a,b], Sebastian Vogel [b], Martin Atzmueller [a,c], Gerard B.M. Heuvelink [d,e]

[a] Joint Lab Artificial Intelligence and Data Science, Osnabrück University, Osnabrück, Germany
[b] Department of Agromechatronics, Leibniz Institute for Agricultural Engineering and Bioeconomy (ATB), Potsdam, Germany
[c] Research Department Cooperative and Autonomous Systems, German Research Center for Artificial Intelligence (DFKI), Osnabrück, Germany
[d] Soil Geography and Landscape Group, Wageningen University and Research, Wageningen, The Netherlands
[e] ISRIC - World Soil Information Wageningen, Wageningen, The Netherlands

ABSTRACT

Machine learning and geostatistics are two fundamentally different frameworks for the prediction and spatial mapping of soil properties. Geostatistics leverages the spatial structure of soil properties, whereas machine learning models capture the relationship between available environmental features and soil properties. We propose a hybrid framework that augments machine learning with spatial context through the engineering of 'spatial lag' features derived from ordinary kriging. We call this approach 'kriging prior regression' (KpR), as it reverses the logic of regression kriging by incorporating kriging outputs before and during the regression step. To evaluate this approach, we assessed both the point and probabilistic prediction performance of KpR, using TabPFN across six field-scale datasets from *LimeSoDa*. These datasets included soil organic carbon, clay content, and pH, along with features derived from remote sensing and in-situ proximal soil sensing. KpR with TabPFN demonstrated reliable uncertainty estimates and accurate predictions in comparison to several other spatial techniques (e.g., regression/residual kriging with TabPFN), as well as to established non-spatial machine learning algorithms (e.g., random forest and categorical boosting). Most notably, it improved the average $R^2$ by ≈30% relative to machine learning algorithms without spatial context. This improvement was due to the strong prediction performance of the TabPFN algorithm itself and the complementary spatial information provided by KpR features. TabPFN is particularly effective for prediction tasks with small sample sizes, common in precision agriculture, whereas KpR can compensate for weak relationships between sensing features and soil properties when proximal soil sensing data are limited. We conclude that KpR with TabPFN is a robust and versatile modelling framework for digital soil mapping in precision agriculture.

## 1. Introduction

Precision agriculture (PA) represents a transformative approach to farming that aims to increase productivity while minimizing environmental harm (Gebbers and Adamchuk, 2010). It does so by adjusting land management practices to the site-specific needs of soils and crops. A central component of PA is the variable-rate application of fertilizers and lime to avoid under- or over-fertilization (Gebbers and Adamchuk, 2010; Viscarra Rossel and Bouma, 2016). For this, farmers require detailed and accurate soil maps with quantified uncertainty to reliably determine the site-specific optimal fertilization rate within a field (Breure et al., 2022). There are two main quantitative approaches to predict and spatially map soil property values: geostatistics and machine learning (ML). Both approaches rely on fundamentally different information to issue their predictions (Heuvelink and Webster, 2022).

Geostatistical models have been used for field-scale soil mapping since the 1980s (Hajrasuliha et al., 1980; Burgess and Webster, 1980), with ordinary kriging (OK) initially being the most commonly applied method (Oliver and Webster, 2015). OK quantifies spatial autocorrelation of a target property by modelling the semivariogram from a set of measured soil samples. It uses this information and the known soil observations to linearly interpolate soil property values to unsampled locations. Under the assumptions of second-order stationarity and a correctly specified semivariogram, OK is an optimal linear interpolator. Lastly, it calculates the prediction error variance at each unsampled location, which serves as an inherent measure of prediction uncertainty (Oliver and Webster, 2015; Heuvelink, 2018).

* Corresponding author at: Joint Lab Artificial Intelligence and Data Science, Osnabrück University, Osnabrück, Germany.
*E-mail address:* jonas.schmidinger@uni-osnabrueck.de (J. Schmidinger).

By the late 1990s and early 2000s, high-resolution soil sensing data became increasingly available through advances in remote sensing (Ge et al., 2011) and the introduction of on-the-go proximal soil sensors (PSSs) (Gebbers and Adamchuk, 2010). This development enabled digital soil mapping (DSM) within a regression framework, where quantitative relationships between target soil properties and features derived from sensors or other environmental data are modelled (McBratney et al., 2003; Scull et al., 2003). By the 2010s, hybrid models that combined a regression model with subsequent kriging of the model residuals, known as regression kriging (RK) or residual kriging (Hengl et al., 2004; Wang et al., 2019), became popular (Chen et al., 2022). However, over the past decade, pure ML methods that treat the task solely as a tabular regression problem have increasingly ousted geostatistics (Chen et al., 2022; Heuvelink and Webster, 2022). Similar trends can likewise be observed across various other spatial mapping disciplines (e.g. Erdogan Erten et al., 2022; Zhan et al., 2023; Cellmer and Kobylińska, 2024).

This shift towards ML is supported by empirical evidence from numerous case studies, which frequently reported that tabular ML outperforms kriging-based approaches in most scenarios (e.g. Tziachris et al., 2020; Safaee et al., 2024). Advantages of geostatistics seem to be limited to specific modelling scenarios, for example when sampling densities are very high (Loiseau et al., 2021), and when only features with insufficient explanatory power, as in forested environments, are accessible (Beguin et al., 2017). Further, a number of case-studies indicated that uncertainty estimates from certain ML techniques were shown to be more reliable than those from RK (Szatmári and Pásztor, 2019; Schmidinger and Heuvelink, 2023). Due to the success of ML, some authors have questioned kriging's relevance in contemporary DSM (Behrens et al., 2024), even though geostatistics retains unique advantages for spatial tasks (Heuvelink and Webster, 2022).

The adoption of ML is only expected to intensify as ML capabilities, computational power, and availability of different high-resolution features continue to grow. One recent example for ML progress is TabPFN, a transformer-based foundation model that employs in-context learning. It is specialized in handling prediction tasks for small- to medium-sized tabular datasets (Hollmann et al., 2025). Small-sized datasets have traditionally been challenging for data-driven ML, limiting applicability in the PA domain, where available training datasets are typically small (Schmidinger et al., 2024b). However, TabPFN has been shown to outperform other ML algorithms on such datasets in various domain-independent ML benchmarking studies (Erickson et al., 2025; Hollmann et al., 2025; Ye et al., 2025) and even in the context of DSM (Barkov et al., 2025).

The undeniable merits and fast-growing capabilities of ML simultaneously led to unrealistic expectations about their applicability (Geirhos et al., 2020). Even the most sophisticated ML techniques are unable to find meaningful patterns when the sensing features lack a substantive relationship with the target soil properties. This is particularly a concern for PA because most remote sensing techniques do not have the required spatial resolution for field-scale applications and PSS data are expensive to produce. Furthermore, most sensing data are only indirectly related to soil properties they are meant to predict. Indirect relationships can be inconsistent in their usefulness, as their validity often depends on specific environmental or management conditions (Gebbers, 2018; Rhymes et al., 2023). For instance, reported correlations between clay content and electrical conductivity range from negligible to strong depending on the study area (e.g. Sudduth et al., 2005; Lukas et al., 2009; Møller et al., 2021). Data fusion from different PSSs usually mitigates this issue (Ji et al., 2019; Schmidinger et al., 2024a; Vullaganti et al., 2025) but is not always economically feasible and requires more training data. In cases where sensing features show little or no relation to a target soil property, OK is an attractive alternative, provided that the target soil property has spatial structure and a sufficiently large dataset is available for the semivariogram estimation (e.g. Pouladi et al., 2019; Bejarano et al., 2025). Hence, integrating geostatistics into ML remains a promising approach for field-scale DSM, because not all sources of spatial variation may be represented by the commonly available sensing features. New spatial techniques beyond RK should be considered for this task.

We propose an alternative to RK that uses the OK prediction and variance as additional spatial features for hybrid modelling. These kriging-derived features can be supplied to any ML algorithm alongside the conventional sensing features used in DSM for PA. We refer to this framework as 'kriging prior regression' (KpR) as it applies kriging *prior* to the regression task, thereby inverting the logic of RK, which uses kriging as a post-processing step. In KpR, the spatial autocorrelation becomes a direct component of the ML function, conceptually similar to autoregressive modelling (Anselin, 1988). In contrast to RK, this can reveal insights into interactions between sensing features and spatial structure, potentially improving predictions.

KpR is conceptually part of 'spatial lag' feature engineering, a concept originating in spatial economics (Anselin, 1988). This technique has largely been disregarded in the context of soil mapping, apart from two notable exceptions (Liu et al., 2022; Chen et al., 2024) and related approaches (Sekulić et al., 2020). In essence, a spatial lag represents the autocorrelation of a variable through a weighted average of neighbouring values, typically computed using $k$-nearest neighbours (KNN) or inverse-distance weighting (IDW) (Li et al., 2017; Liu et al., 2022; Soltani et al., 2022; Chen et al., 2024). To our knowledge, a spatial lag feature based on (ordinary) kriging, as proposed with KpR, has not yet been used. Yet OK, as a geostatistical method, should capture the spatial structure patterns more effectively than deterministic interpolation techniques like IDW (Goovaerts, 2000). Although they share the same name, 'spatial lag' in feature engineering is unrelated to the 'spatial lag' used in variogram analysis. To avoid confusion, we adopt the more methodological term 'interpolation prior regression' for this type of feature engineering, following the naming convention of the proposed KpR (i.e., 'kriging prior regression') approach.

We assume that the combination of TabPFN and KpR constitutes a modelling framework more suitable for common field-scale DSM tasks in PA, given the aforementioned advantages of both methods. We test this hypothesis by evaluating whether this framework improves the predictive accuracy compared to established techniques across different field-scale datasets and soil properties. To provide broader context, we benchmark this approach with multiple other spatial techniques and learning algorithms. However, for the main analysis we restrict our comparison primarily to TabPFN with RK and KpR, as well as two baseline models: TabPFN without spatial context and OK. Given the importance of accounting for prediction uncertainty in agronomic decision-making (Breure et al., 2022; Lark et al., 2022; Takoutsing et al., 2025), we also evaluate their uncertainty estimates. Finally, our analysis explicitly evaluates the performance in dependence on dataset and model characteristics to identify modelling conditions in which TabPFN may benefit most from the spatial context of KpR.

## 2. Materials and methods

### 2.1. Study design

The main workflow for this study is shown in Fig. 1. Six datasets with soil organic carbon (SOC), clay and pH as target soil properties together with various sensing features were included (Section 2.2). Therefore, the study consisted of 18 prediction tasks in total. In the main analysis, we compared four prediction frameworks across these tasks in terms of their point prediction performance and uncertainty estimates. These prediction frameworks were: (1) OK, (2) TabPFN with common sensing features, hereafter referred to as (baseline) TabPFN, (3) TabPFN with additional KpR features, hereafter referred to as TabPFN-KpR, and (4) TabPFN with adjustment of the predictions through kriging

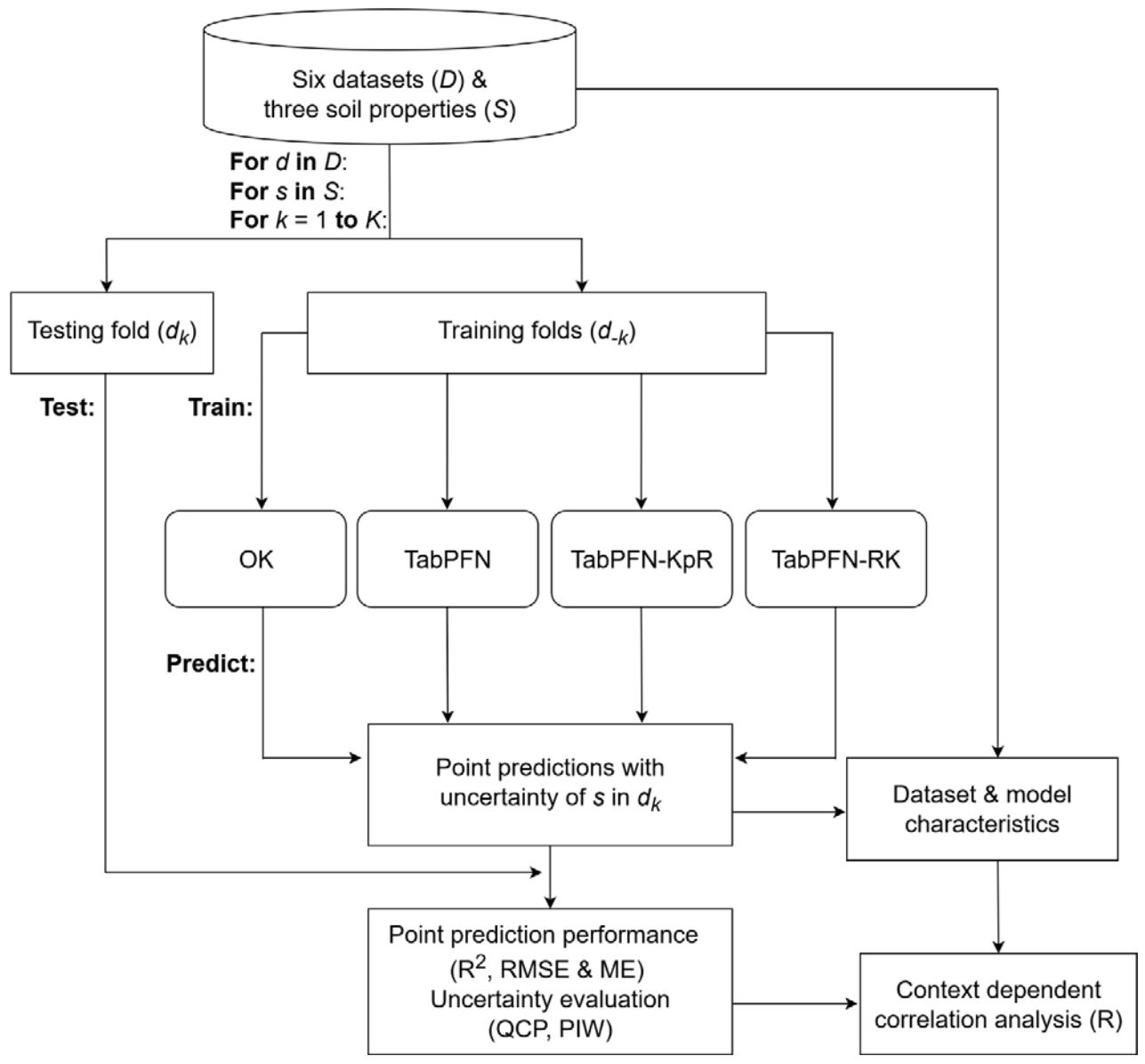

**Fig. 1.** Workflow of the statistical analysis presented in the main results section. Further methods tested and listed in the Appendix are not shown in this figure.

of the model residuals, i.e., RK, hereafter referred to as TabPFN-RK (Section 2.3).

Beyond the main analysis (Fig. 1), we conducted a broader analysis presented in the Appendix, where we extended the TabPFN spatial-techniques comparison to include eight additional methods (Appendix A). These spatial methods include: (1) geographic coordinates as features, (2) Euclidean distance fields (Behrens et al., 2018), (3) model stacking with OK (Erdogan Erten et al., 2022), (4) KNN distance and observation matrix features (Sekulić et al., 2020), (5) IDW prior regression (Chen et al., 2024), (6) KNN prior regression (Liu et al., 2022), (7) leave-one-out (LOO) RK on independent residuals, and (8) a reduced KpR framework. This benchmark, therefore, covers the majority of techniques suggested in the spatial ML literature (Jemeljanova et al., 2024).

Apart from the different spatial prediction methods, we also compared TabPFN-KpR with non-spatial ML algorithms commonly used in DSM to contextualize the obtained performance with standard practice. For that, we used the results obtained in Schmidinger et al. (2025), a previous benchmarking study on the same datasets involving four additional regression techniques: random forest, categorical boosting (CatBoost), multiple linear regression, and support vector regression. The same validation strategy was used to ensure direct comparability. Hence, all models were evaluated using random $K$-fold cross-validation (CV) with $K = 10$, applying the same CV splits as in Schmidinger et al. (2025). Point prediction performances were compared based on the coefficient of determination ($R^2$), while supplementary results in Appendix B are presented with the root-mean-square error (RMSE) and mean error (ME). Uncertainty estimates were compared based on the quantile coverage probability (QCP) and prediction interval width (PIW) (Section 2.4).

The ability of a method to make accurate predictions can change depending on the nature of a dataset and prediction task (Nießl et al., 2022). To understand how contextual differences affect the model performance, we did an exploratory analysis by correlating dataset and model characteristics with the $R^2$ or difference in $R^2$ ($\Delta R^2$) between methods. The evaluated characteristics are briefly listed in Section 2.5.

### 2.2. LimeSoDa datasets

The datasets used in the statistical analysis were sourced from the Precision Liming Soil Datasets (LimeSoDa) benchmarking collection (Schmidinger et al., 2025). LimeSoDa is an open dataset collection with a variety of different field-scale datasets for the benchmarking and testing of statistical methods in the context of soil mapping for precision liming. SOC, pH and clay content are the target soil properties in each dataset but the sensing features are dataset-specific.

Overall, LimeSoDa consists of 31 datasets. Since not all datasets of LimeSoDa were relevant for the scope of this study, we selected them based on three criteria: (1) A dataset had to contain features from at least one on-the-go PSS to evaluate datasets most realistic for reliable field-specific PA, (2) the reference sampling needed to be based on a non-targeted and non-spatially clustered sampling to allow for straightforward validation and interpretation of the map quality (Wadoux et al., 2021), and (3) a dataset was required to contain more than 50 soil samples, as the semivariogram may not be adequately estimated with fewer observations (Webster and Oliver, 1992). This resulted in six datasets: two from Brazil and four from Germany. The dataset IDs are BB.250, G.150, MG.112, SA.112, BB.72, and RP.62. An overview of the datasets is given in Table 1. Note that the overall number of sensors is not equal to the number of PSSs in Table 1, because some sensing information was obtained from satellites, i.e., remote sensing. The soil sampling configurations of the six datasets are shown in Fig. 2. SA.112 and RP.62 were published without geographic coordinates for privacy protection, but we obtained permission to use the geographic coordinates for our analysis. More detailed information about the soil analysis and sensing campaigns can be found in the metadata of the LimeSoDa R and Python packages.

For SA.112 and RP.62, we applied principal component analysis on the high-dimensional NIR features and used the first ten components for the modelling, without processing the remaining sensing features. In prior research, the high feature dimensionality of NIR proved to be detrimental for common learning algorithms unless reduced (Schmidinger et al., 2025). No further pre-processing was conducted for any other features.

### 2.3. Prediction models

#### 2.3.1. Ordinary kriging (OK)

OK is a geostatistical interpolation method (Oliver and Webster, 2015). It leverages the spatial autocorrelation from measured soil samples $y(s_i)$ at sampled locations $s_i$ ($i = 1, 2, \dots, n$) to make predictions $\hat{y}_{OK}(s_0)$ for any unsampled location $s_0$ in the study area of interest. In contrast to a ML approach, it is purely based on the spatial dependence of the $y(s_i)$. Hence, $\hat{y}_{OK}(s_0)$ is a weighted mean of the $y(s_i)$:

$$\hat{y}_{OK}(s_0) = \sum_{i=1}^{n} \lambda_i \, y(s_i), \tag{1}$$

where the $\lambda_i$ are the specific kriging weights and $n$ is the number of soil samples available for training. The $\lambda_i$ are chosen to minimize the expected squared prediction error, as assessed through the kriging variance $\sigma^2_{OK}$, subject to the unbiasedness constraint that $\sum_{i=1}^{n} \lambda_i = 1$. Solving for $\lambda_i$ requires information on the spatial autocorrelation, which is quantified by the semivariogram $\gamma(h)$, defined as:

$$\gamma(h) = \frac{1}{2} E[(Y(s) - Y(s+h))^2], \tag{2}$$

where $E$ is the mathematical expected value, $h$ is the geographic distance between two locations and $Y$ a postulated geostatistical model of the soil property of interest. Using the semivariogram, the OK system can be solved for $\lambda_i$. Additionally, $\sigma^2_{OK}(s_0)$ quantifies the prediction uncertainty of $\hat{y}_{OK}(s_0)$ through the expected error variance (Oliver and Webster, 2015):

$$\sigma^2_{OK}(s_0) = \sum_{i=1}^{n} \lambda_i \gamma(s_i - s_0) + \psi, \tag{3}$$

**Table 1**
Overview for the selected datasets.

| Dataset ID | Location | Size of study area (ha) | Sample size | Sensors[a] | Number of PSSs | Sampling design |
|---|---|---|---|---|---|---|
| BB.250 | Brandenburg, Germany | 52 | 250 | DEM, ERa, Gamma, pH-ISE, RSS, VI | 3 | Triangular grid sampling |
| G.150 | Goias, Brazil | 79 | 150 | DEM, ERa, RSS, VI | 1 | Regular grid sampling |
| MG.112 | Mato Grosso, Brazil | 111 | 112 | DEM, ERa, RSS, VI | 1 | Regular grid sampling |
| SA.112 | Saxony-Anhalt, Germany | 27 | 112 | DEM, ERa, Gamma, NIR, pH-ISE, VI | 4 | Incomplete regular grid sampling |
| BB.72 | Brandenburg, Germany | 3.4 | 72 | DEM, ERa, Gamma, pH-ISE, RSS, VI | 3 | Triangular grid sampling |
| RP.62 | Rhineland-Palatinate, Germany | 3.3 | 62 | ERa, Gamma, NIR, pH-ISE, VI | 4 | Regular grid sampling |

[a] Abbreviations: Digital elevation model and terrain parameters (DEM); Apparent electrical resistivity (ERa); Gamma-ray activity (Gamma); Near infrared spectroscopy (NIR); Ion selective electrodes for pH determination (pH-ISE); Remote sensing derived spectral data (RSS); Vegetation indices (VI).

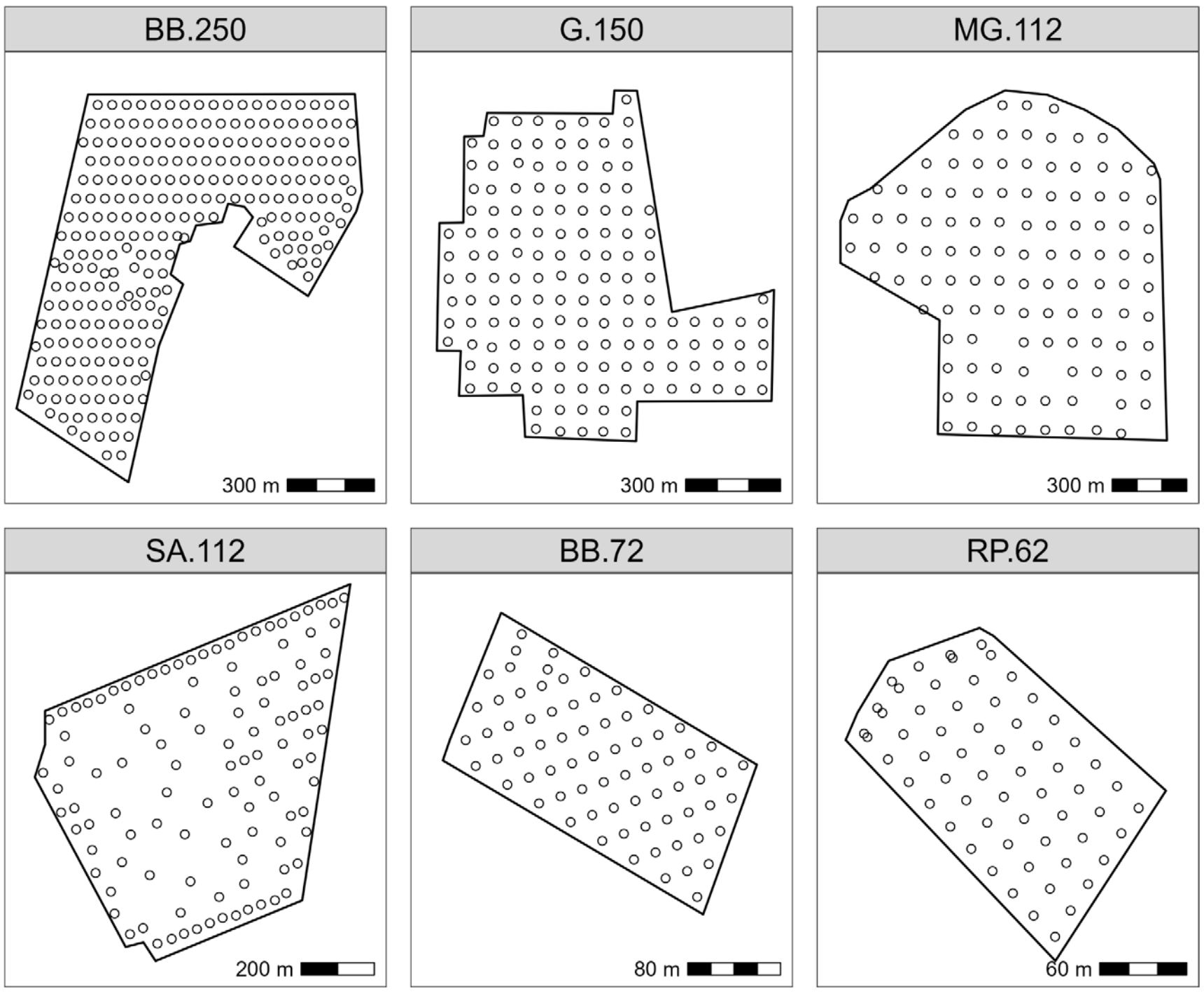

**Fig. 2.** Soil sample configurations of the selected datasets from LimeSoDa.

where $\psi$ is a Lagrange parameter.

Throughout the analysis we used *automap* (Hiemstra and Hiemstra, 2013), which automatizes the semivariogram model fitting of *gstat* (Pebesma, 2006).

### 2.3.2. TabPFN

TabPFN, short for tabular prior-data fitted network, is a foundation model designed for small- to medium-sized tabular prediction tasks (Hollmann et al., 2023, 2025). It is a transformer-based neural network pre-trained on millions of synthetic datasets. These synthetic datasets were constructed to imitate how, in real-world datasets, certain features relate to the target properties. They were generated through structural causal models with different underlying parameters. Once pre-trained, TabPFN can leverage in-context learning to issue predictions for any given real-world data input. TabPFN approximates Bayesian inferences by creating the posterior predictive distribution $p(y(s_0) \mid X(s_0), d_{\text{train}})$ using the prior distribution defined by the synthetic data (Müller et al., 2022), where $X$ are the sensing features and $d_{\text{train}}$ the 'context' of the real data provided to TabPFN. As is typical in Bayesian frameworks, this posterior predictive distribution aims to reflect the underlying prediction uncertainty.

In its original release (Hollmann et al., 2023), TabPFN was limited to classification tasks. The current version (Hollmann et al., 2025), sometimes referred to as TabPFN v2 in the literature, extends support to regression tasks, making it more applicable for DSM. We refer to this latest version simply as TabPFN throughout this paper.

A key characteristic of TabPFN is that it does not require conventional dataset-specific hyperparameter tuning. As demonstrated in Hollmann et al. (2025) or the current results of the TabArena scoreboard (accessed on 10 September 2025) (Erickson et al., 2025), TabPFN can achieve highly competitive performances using default settings for small datasets. For simplicity, we therefore implemented TabPFN with its default configuration in our experiments.

### 2.3.3. TabPFN with kriging prior regression (TabPFN-KpR)

Interpolation prior regression (IpR) (i.e., spatial lag feature engineering) is a framework to generate spatial features based on neighbouring observations (Liu et al., 2022). Traditionally, it has been created through a weighted mean of a preliminary spatial interpolation $\hat{y}_{\text{IpR}}(s_0)$:

$$\hat{y}_{\text{IpR}}(s_0) = \sum_{i=1}^{n} w_i y(s_i), \tag{4}$$

where the $w_i$ are weights. These weights may be obtained through simple KNN averaging prior regression, i.e., KNNpR (Liu et al., 2022), or IDW prior regression, i.e., IDWpR (Chen et al., 2024). Similarly, OK is also a weighted mean as established in Section 2.3.1. In principle non-linear interpolation methods could also be used within IpR (e.g. Chen et al., 2020), but, to our knowledge, this has not yet been done.

$\hat{y}_{\mathrm{IpR}}(s_0)$ is then appended to $X(s_0)$, to encode information on the spatial autocorrelation of $y$ as additional feature:

$$X_{\mathrm{IpR}}(s_0) = \{X(s_0),\ \hat{y}_{\mathrm{IpR}}(s_0)\}, \tag{5}$$

where $X_{\mathrm{IpR}}(s_0)$ is the new augmented feature space.

In our proposed KpR variant, we extend $X(s_0)$ with $\hat{y}_{\mathrm{OK}}(s_0)$ and $\sigma^2_{\mathrm{OK}}(s_0)$ as an additional feature:

$$X_{\mathrm{KpR}}(s_0) = \{X(s_0),\ \hat{y}_{\mathrm{OK}}(s_0),\ \sigma^2_{\mathrm{OK}}(s_0)\}. \tag{6}$$

Adding $\sigma^2_{\mathrm{OK}}(s_0)$ could be useful because it can assist the ML model to determine how much weight to place on $\hat{y}_{\mathrm{OK}}$. For example, if the separation distances between the observations and a prediction location are small, $\sigma^2_{\mathrm{OK}}$ will be small as well and thus $\hat{y}_{\mathrm{OK}}$ may tend to be more informative than for prediction locations far away from observation locations. This could become relevant in the case of spatially clustered datasets.

We used a LOO scheme to create $\hat{y}_{\mathrm{OK}}(s_i)$ and $\sigma^2_{\mathrm{OK}}(s_i)$ for the training dataset without introducing data leakage. In other words, we excluded $y(s_i)$ for the calculation of $\hat{y}_{\mathrm{OK}}(s_i)$. This prevents a scenario where $\hat{y}_{\mathrm{OK}}(s_i) = y(s_i)$, which would lead to pure overfitting on $\hat{y}_{\mathrm{OK}}(s_i)$.

IpR is sometimes mistakenly believed to cause data leakage by default (Ohmer et al., 2025). To untangle the root of this misconception, we have included a section about data leakage in Appendix C.

The R-code for KpR is published as a generic function in a GitHub tutorial at https://jonasschmidinger.github.io/Kriging_prior_Regression/.

### 2.3.4. TabPFN regression kriging (TabPFN-RK)

RK has a long-standing history in soil mapping in combination with linear models (Hengl et al., 2004). RK in combination with ML, sometimes referred to as residual kriging (Wang et al., 2019), has been less frequently used, but follows the same principles. RK makes use of the residuals $e$ from the training data of a regression model, for which we use TabPFN:

$$e_{\mathrm{TabPFN}}(s_i) = y(s_i) - \hat{y}_{\mathrm{TabPFN}}(X(s_i)), \tag{7}$$

OK is then performed on $e_{\mathrm{TabPFN}}$ to predict the residuals at unobserved locations:

$$\hat{e}_{\mathrm{TabPFN}}(s_0) = \sum_{i=1}^{n} \lambda_i\, e_{\mathrm{TabPFN}}(s_i). \tag{8}$$

where the $\lambda_i$ are now kriging weights derived from the residual semivariogram. Hence, RK is expected to improve predictions in the case of spatial autocorrelation for $e_{\mathrm{TabPFN}}$.

Finally, the prediction $\hat{y}_{RK}(s_0)$ is obtained by summing $\hat{y}_{\mathrm{TabPFN}}(s_0)$ and $\hat{e}_{\mathrm{TabPFN}}(s_0)$:

$$\hat{y}_{\mathrm{RK}}(s_0) = \ \hat{y}_{\mathrm{TabPFN}}(X(s_0)) + \hat{e}_{\ \mathrm{TabPFN}}(s_0). \tag{9}$$

The residual kriging variance $\sigma^2_{\mathrm{RK}}(s_0)$ corresponding to $\hat{e}_{\mathrm{TabPFN}}(s_0)$ can then be used as the uncertainty estimate of RK. This differentiates RK from KpR, as RK determines the prediction uncertainty directly through geostatistics, while in TabPFN-KpR it is derived from the posterior distribution of the Bayesian approximation $p(y(s_0) \mid X_{\mathrm{KpR}}(s_0), d_{\mathrm{train}})$ (see Section 2.3.3).

This standard implementation of RK calculates residuals with a trend fitted on the same observations used to fit the trend, which is a form of data leakage. Similar to the KpR framework, we therefore also introduce and compare a RK framework based on inner-loop residuals from a nested LOO CV to avoid data leakage in Appendix A. This leakage-free variant applies kriging on independent residuals from LOO CV.

## 2.4. Model evaluation

### 2.4.1. Point predictions

ME, RMSE and $R^2$ were determined from predictions and independent observations using 10-fold CV. They were calculated as:

$$\mathrm{ME} = \frac{1}{n}\sum_{i=1}^{n}\left(y_i - \hat{y}_i\right), \tag{10}$$

$$\mathrm{RMSE} = \sqrt{\frac{1}{n}\sum_{i=1}^{n}\left(y_i - \hat{y}_i\right)^2}, \tag{11}$$

$$\mathrm{R}^2 = 1 - \frac{\sum_{i=1}^{n}\left(y_i - \hat{y}_i\right)^2}{\sum_{i=1}^{n}\left(y_i - \bar{y}\right)^2}, \tag{12}$$

where $\bar{y}$ is the mean of the observations.

$R^2$ was calculated with the same formula as that of the Nash–Sutcliffe model efficiency coefficient (Nash and Sutcliffe, 1970).

### 2.4.2. Uncertainty estimates

The coverage of the quantiles (i.e., QCP) was used to evaluate the reliability of uncertainty estimates (Pinson et al., 2007). Conceptually, QCP has the same logic as the more commonly known prediction interval coverage probability but QCP evaluates quantiles to account for a one-sided bias (Pinson et al., 2007; Schmidinger and Heuvelink, 2023). It measures how frequently the true values $y_i$ fall below the associated predictive quantiles $\hat{q}_i$ across a dataset:

$$\mathrm{QCP}(\tau) = \frac{1}{n}\sum_{i=1}^{n}\mathbb{1}\left(y_i \leq \hat{q}_i^{\tau}\right), \tag{13}$$

where $\tau$ is the evaluated quantile level and $\mathbb{1}$ is an indicator function defining whether the observation is below or above the predicted quantile value. For $\tau$ we evaluated:

$$\tau \in \{0.001,\ 0.1,\ 0.2,\ \ldots,\ 0.9,\ 0.999\}. \tag{14}$$

QCP was evaluated in the form of reliability plots and two summary measures from all prediction tasks: the mean absolute deviation, $\overline{\delta}_{\mathrm{QCP}}$, and the median absolute deviation, $\widetilde{\delta}_{\mathrm{QCP}}$. For a single dataset, $\delta_{\mathrm{QCP}}$ was defined as $|\mathrm{QCP}(\tau) - \tau|$, averaged over all elements of $\tau$.

The sharpness of the uncertainty estimates was measured by the width of the prediction intervals (i.e., PIW) (Pinson et al., 2007). While strictly speaking it is not a validation method, it helps interpreting the results. It is calculated as:

$$\mathrm{PIW}(\alpha) = \frac{1}{n}\sum_{i=1}^{n}\left(q_{i,1-\frac{\alpha}{2}} - q_{i,\frac{\alpha}{2}}\right), \tag{15}$$

where $\alpha$ is the target error rate of the prediction interval. It was calculated for all pairs of $\tau$ that can form a central interval (e.g., 0.1 and 0.9 form the 80% interval, 0.2 and 0.8, the 60% interval etc.). Further, PIW was min–max normalized per prediction task across the model outcomes to be unit independent.

## 2.5. Dataset and model characteristics

We evaluated contexts in which TabPFN-KpR outperformed or underperformed relative to OK and, in particular, the baseline TabPFN. To achieve this, we analysed the correlation of several dataset characteristics with the predictive performance ($R^2$) across the prediction tasks. Further, the difference between $R^2_{\text{TabPFN-KpR}}$ and $R^2_{\text{TabPFN}}$ ($\Delta_{\text{TabK}-\text{Tab}}R^2$) was measured to compute its correlation with several dataset and model characteristics. We prioritized linear relationships in this analysis; therefore, the strength of association was quantified using Pearson's correlation coefficient (R), and statistical significance was assessed via $p$-values derived from a $t$-test. In the main results section, we will present only those characteristics which showed at least some relation to the $\Delta_{\text{TabK}-\text{Tab}}R^2$, using a cutoff of $p < 0.2$. We have used a relatively relaxed $p$ threshold, to present a broader set of potentially relevant characteristics, given the limited statistical power associated with 18 prediction tasks. In Appendix B, we present all remaining results of the correlation analyses.

As dataset characteristics, we considered seven properties: (1) sample size, (2) number of PSSs (Table 2), (3) spatial autocorrelation of the target soil property, (4) variance of the local spatial autocorrelation, (5) frequency of spatial outliers, (6) normalized standard deviation (SD) of the soil properties, and (7) skewness of the target soil property values. The spatial autocorrelation of the target soil properties was quantified through the global Moran's I (MI), using the six nearest neighbours and IDW weights via the spdep framework (Pebesma and Bivand, 2023). For simplicity, we did not account for field size or sampling density in the MI calculations. The variance of the local MI (Anselin, 1995) (i.e., local spatial autocorrelation) was used as an indicator of spatial stationarity, and to determine the frequency of spatial outliers. Observations were classified as spatial outliers if their Local MI was statistically significant ($\alpha = 0.05$) and indicated an anomaly through a high target value surrounded by low values, or vice versa. Lastly, we included the SD per soil property, which was calculated for each dataset and soil property but was in the end min–max normalized by the soil property, to be unit-independent. The skewness was measured by the adjusted Fisher-Pearson standardized moment coefficient (Doane and Seward, 2011).

As model characteristics, we defined five different properties and investigated their relation to $\Delta_{\mathrm{TabK-Tab}}\mathrm{R}^2$. This included: (1) spatial autocorrelation of the test-fold errors for OK, (2) spatial autocorrelation of the test-fold errors for the baseline TabPFN, (3) difference in performance between TabPFN and OK ($\Delta_{\mathrm{Tab-OK}}\mathrm{R}^2$), (4) absolute performance of OK ($\mathrm{R}^2_{\mathrm{OK}}$), and (5) absolute performance of TabPFN ($\mathrm{R}^2_{\mathrm{TabPFN}}$).

## 3. Results and discussions

### 3.1. General comparison

#### 3.1.1. Point predictions

TabPFN outperformed OK on average, achieving a higher mean $R^2$ of 0.56 compared to OK's mean $R^2$ of 0.50. Although this supports the prevailing notion that ML usually outperforms kriging, the performance gap was not overwhelmingly large. In fact, in four prediction tasks OK produced better predictions than TabPFN, in some cases by a substantial margin. The most significant example was SOC in G.150, where OK achieved an $R^2$ of 0.60, clearly outperforming the 0.35 of TabPFN (Table B.1).

TabPFN-RK and TabPFN-KpR achieved a mean $R^2$ of 0.58 and 0.60, respectively. Therefore, ML combined with geostatistical methods led to higher prediction accuracy than either ML or geostatistics alone. The proposed TabPFN-KpR was on average the most accurate framework in this main analysis, with a slight margin over TabPFN-RK. Even in comparison to eight further spatial techniques combined with TabPFN (Appendix A, Table A.1), TabPFN-KpR remained a top performer. Only TabPFN paired with IDWpR and TabPFN-KpR without $\sigma^2$ had about the same prediction performance, with marginally higher mean $R^2$ values but lower median $R^2$ values. Despite a similar empirical performance, TabPFN-KpR has advantages over these methods that extend beyond this study, which we further discuss in Appendix A.

Given just a few other benchmarking studies on spatial ML techniques (e.g. Beguin et al., 2017; Milà et al., 2024; Kmoch et al., 2025), this is, to our knowledge the most comprehensive comparison and introduces KpR as a new capable spatial feature engineering technique. However, we encourage further evaluation of its applicability beyond the field-scale domain.

To contextualize these findings with the previous initial LimeSoDa benchmarking study (Schmidinger et al., 2025), Table 2 presents the results of other ML models evaluated on the same prediction tasks. It shows that TabPFN, especially when combined with KpR, substantially outperforms them. Compared to the best selected non-spatial ML algorithm, TabPFN-KpR increased the $R^2$ by 18% from 0.51 (CatBoost) to 0.60 and the median $R^2$ by 22% from 0.49 (Support Vector Regression) to 0.60. Compared to all four methods, it on average improved the $R^2$ by even ≈30%. This considerable improvement is attributable to the inherent effectiveness of the TabPFN algorithm and complementary information from KpR features. It demonstrates an untapped potential to improve DSM in PA, purely through more appropriate and novel statistical solutions.

**Table 2**
Summary statistics of $R^2$ aggregated over all prediction tasks; extended with results from a former benchmarking study on the same prediction tasks. The full table with results for each individual prediction task is shown in Appendix B; Table B.1.

| Method | Mean $R^2$ | Median $R^2$ | SD $R^2$ |
|---|---|---|---|
| OK | 0.501 | 0.540 | 0.203 |
| TabPFN | 0.560 | 0.576 | 0.231 |
| TabPFN-RK | 0.585 | 0.582 | 0.209 |
| TabPFN-KpR | **0.597** | **0.598** | 0.213 |
| Multiple linear regression[a] | 0.328 | 0.402 | 0.345 |
| Support vector regression[a] | 0.463 | 0.487 | 0.264 |
| Random forest[a] | 0.502 | 0.484 | 0.254 |
| CatBoost[a] | 0.507 | 0.476 | 0.246 |

[a] For more information about the methodology and tuning see Schmidinger et al. (2025).

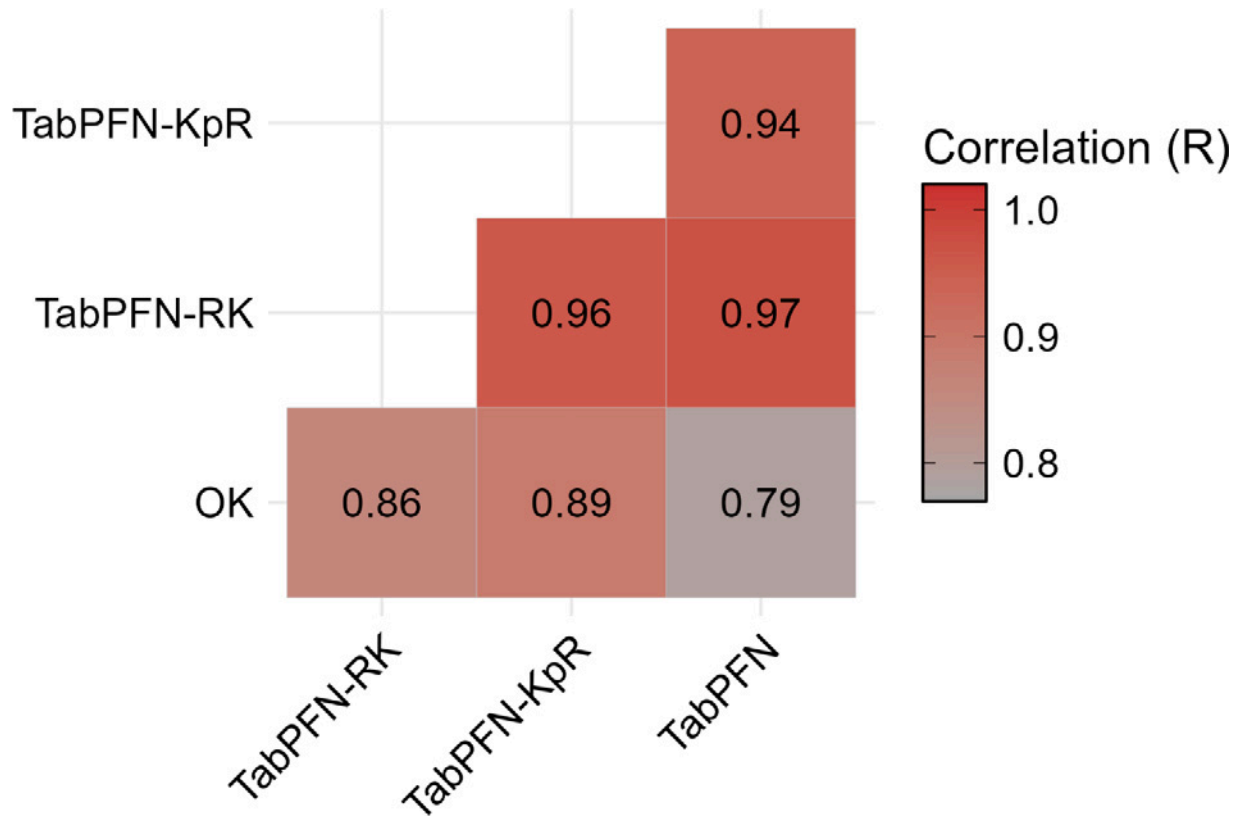

**Fig. 3.** Pearson correlations between predictions of the different methods averaged across all prediction tasks.

Barkov et al. (2025) already demonstrated a strong performance of TabPFN across the full LimeSoDa collection, in line with several other recent ML benchmarks (Erickson et al., 2025; Hollmann et al., 2025; Ye et al., 2025). Its strong performance for small datasets is an important aspect in the context of PA because the commonly small amount of training data is used to constrain ML applicability for field-scale DSM (Schmidinger et al., 2024b; Bejarano et al., 2025).

The addition of KpR features further improved the accuracy by enabling TabPFN to draw on the spatial information extracted by OK. This is reflected in the correlation plot in Fig. 3. It shows that predictions issued by TabPFN-KpR became more correlated to those of OK with an r of 0.89, while the TabPFN baseline was less correlated with OK (R = 0.79). Although a similar trend can be observed for TabPFN-RK, its predictions were less correlated with OK (R = 0.86). This correlation pattern suggests that TabPFN-KpR has relied more on spatial information from kriging than TabPFN-RK, which has proven advantageous for the prediction tasks evaluated. These findings also suggest that geostatistics remains relevant for DSM even with the emergence of strong ML models like TabPFN. Although OK remains limited as a traditional stand-alone prediction model compared to ML, it may thrive in new supporting roles such as KpR.

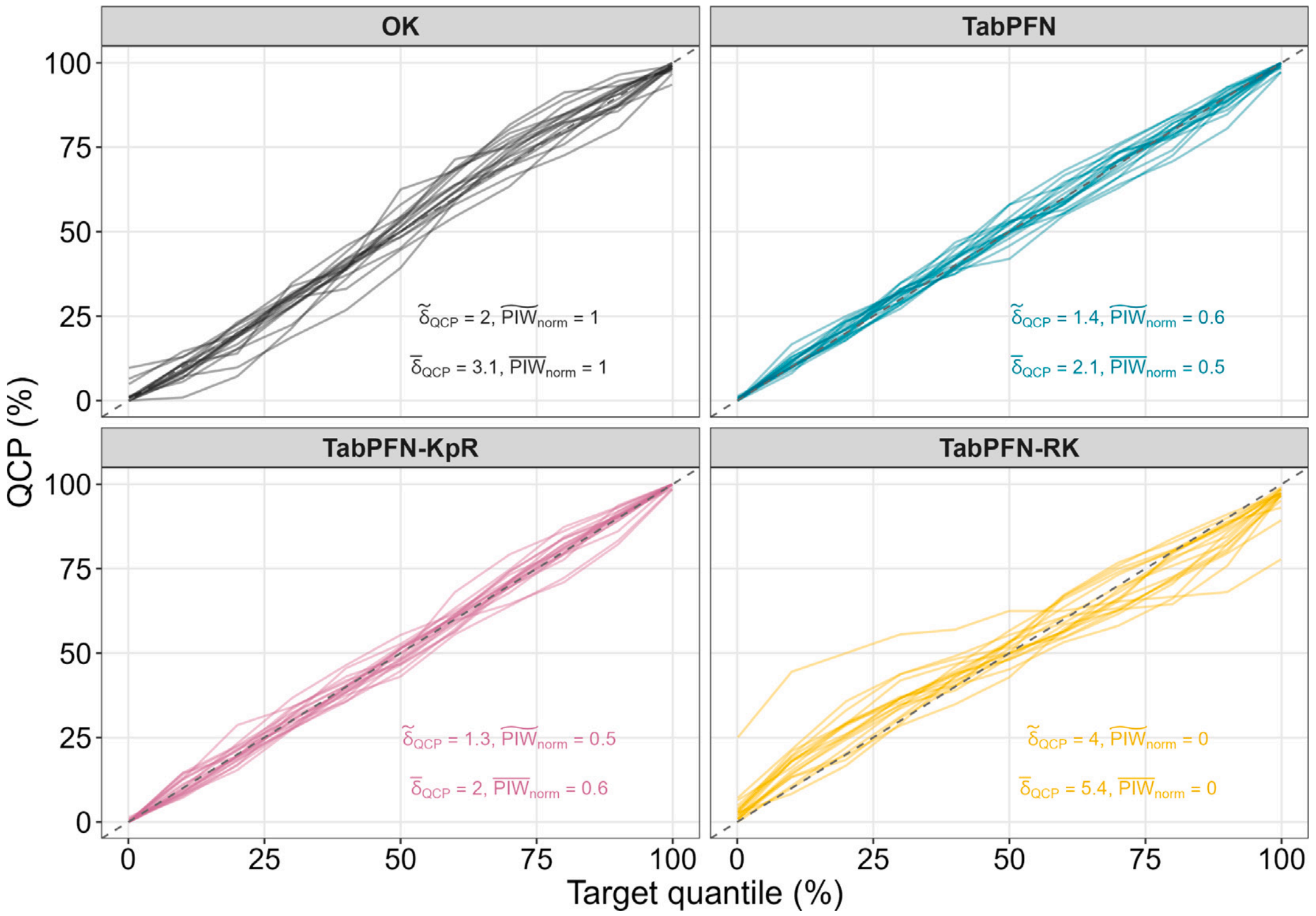

**Fig. 4.** Reliability plots of the quantile coverage with the different prediction methods. Each line represents the QCP obtained from one of 18 prediction tasks. Lines closer to the 1:1 line indicate more reliable uncertainty estimates. The mean and median deviation from the 1:1 line is measured by $\overline{\delta}_{QCP}$ and $\tilde{\delta}_{QCP}$, and the mean and median normalized PIW by $\overline{PIW}_{norm}$ and $\widetilde{PIW}_{norm}$.

#### 3.1.2. *Uncertainty estimates*

As shown in the QCP reliability plots (Fig. 4) for TabPFN and TabPFN-KpR, the TabPFN algorithm was able to issue reliable uncertainty estimates across most prediction tasks, as it had a low $\overline{\delta}_{QCP}$ of about only ∼2. It achieved these reliable uncertainty estimates in one go through its Bayesian approximation. This constitutes an important advantage over other ML methods that often rely on post-processing such as conformal(ised) predictions to ensure reliable uncertainty estimates (Kakhani et al., 2024; Huang et al., 2025), requiring a separate calibration step. However, given that training data is limited in PA, we ideally do not want to limit our training sample size further (Barkov et al., 2024) or have to rely on computationally expensive frameworks (Barber et al., 2021). The capability of reliable probabilistic predictions was usually not the focus in previous benchmarks with TabPFN (e.g. Barkov et al., 2025; Erickson et al., 2025; Ye et al., 2025) but given the critical role of uncertainty assessment in agronomic decision-making (Breure et al., 2022; Lark et al., 2022; Takoutsing et al., 2025), its reliable and straightforward probabilistic predictions further underscore a benefit for PA.

The uncertainty estimates quantified through geostatistics were less reliable. While OK still offered decent estimates for most prediction tasks, the reliability of TabPFN-RK fluctuated greatly with an $\overline{\delta}_{QCP}$ of ∼5. The 1:1 line in Fig. 4 shows a strong overoptimistic tendency, with lower quantiles placed too high and upper quantiles too low, yielding intervals that are too narrow (Schmidinger and Heuvelink, 2023). This is further reinforced by the fact that TabPFN-RK generally provided an $\overline{PIW}_{norm}$ of 0. This means that the PIWs of RK were the narrowest across all prediction tasks, resulting in overoptimistic uncertainty estimates. Similar concerns regarding the reliability of uncertainty estimates of RK compared to other ML approaches have previously been raised in case studies (Szatmári and Pásztor, 2019; Schmidinger and Heuvelink, 2023), but we now demonstrate that this is a recurrent, potentially systematic, issue for RK. We attribute RK's consistently overoptimistic uncertainty estimates to its use of residuals that were computed by subtracting ML predictions from training observations, where the training observation was used to derive the prediction. This led to a form of data leakage, as the ML predictions are not independent from the training observations, resulting in an underestimated residual variance. We discuss this flaw of RK further in Appendix B.3 and demonstrate how an adjusted LOO RK approach with independent residuals avoids the leakage problem. We found that LOO RK returns similar acceptable QCP values as OK. This overoptimistic tendency of the classical RK implementation would inevitably lead to unaccountable adverse effects such as under- or over-fertilization if used in agronomic decisions as in Breure et al. (2022).

In summary, TabPFN-KpR provided the most accurate point predictions and reliable uncertainty estimates. In contrast, the commonly used RK approach for hybrid modelling issued slightly worse point predictions but had considerable deficits in the uncertainty estimation. These results suggest that KpR has significant advantages over RK. In the subsequent sections, we therefore exclude TabPFN-RK from our analysis and focus on the contextual performance of TabPFN-KpR against the baseline TabPFN and OK.

### 3.2. *Contextual performance*

Although TabPFN-KpR led on average to more accurate predictions, it did not enhance the performance of TabPFN for each and every prediction task. Fig. 5 shows the $R^2$ of OK, TabPFN-KpR, and TabPFN for the same prediction tasks connected through a line to highlight the best method per prediction task. For half of the prediction tasks, adding OK features was insignificant for the model performance. For pH predictions in dataset RP.62, using TabPFN-KpR was even highly detrimental, as the $R^2$ decreased by 0.14 compared to the TabPFN baseline (Table B.1). This is not unexpected because no statistical technique is guaranteed to perform best for each and every situation (Nießl et al., 2022). Additional features can foster overfitting if they do not provide complementary information to the learning algorithm (Hughes, 1968). As a consequence, an occasional decrease in accuracy can be observed with TabPFN-KpR compared to the baseline TabPFN. We sought to understand under which dataset conditions the addition of KpR features in TabPFN proved advantageous and when it failed to provide improvements. We left out OK from this explicit comparison as it proved to be best in only one case (Fig. 5).

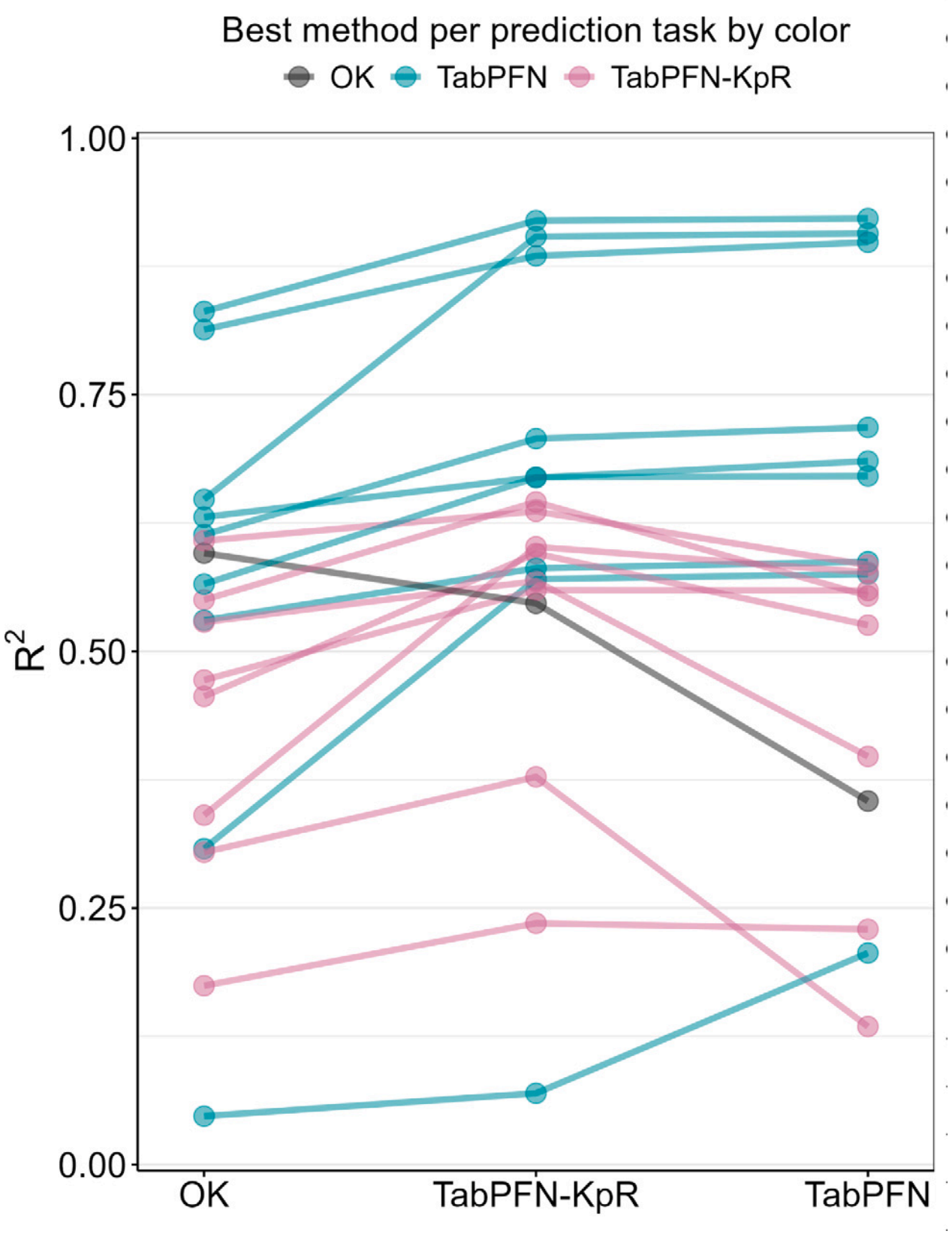

**Fig. 5.** $R^2$ performance across all 18 prediction tasks with three prediction methods. For each task, points represent the $R^2$ achieved, with lines connecting the same task across the methods. The colour indicates the best-performing method for each prediction task.

### 3.2.1. Dataset characteristics

Fig. 6a, c, e illustrate the relation between the $R^2$ of TabPFN-KpR, TabPFN and OK with certain dataset characteristics. Fig. 6b, d, f display the same characteristics, but in relation to $\Delta_{\text{TabK-Tab}}R^2$; defined as the difference between $R^2_{\text{TabPFN-KpR}}$ and $R^2_{\text{TabPFN}}$. A positive $\Delta_{\text{TabK-Tab}}R^2$ reflects an improvement with TabPFN-KpR over the baseline TabPFN, while a negative $\Delta_{\text{TabK-Tab}}R^2$ reveals the opposite. Fig. 6 focuses on dataset characteristics most strongly associated with $\Delta_{\text{TabK-Tab}}R^2$ but further characteristics are presented in Appendix B, Fig. B.1. Overall, there were several dataset characteristics strongly correlated with the absolute $R^2$ of the different models, whereas only a few characteristics had at most a moderate relation with the $\Delta_{\text{TabK-Tab}}R^2$ (Fig. 6 & Fig. B.1).

A higher degree in spatial autocorrelation of the target soil properties, as quantified by the $MI_y$, was strongly positively correlated with the $R^2$, regardless of the modelling technique (Fig. 6a). This outcome is as expected because a high $MI_y$ indicates clear spatial continuity and less noise, making prediction tasks more straightforward. Somewhat unexpectedly, $MI_y$ was about as strongly correlated with OK as with TabPFN. Given that OK explicitly relies on the spatial autocorrelation, whereas TabPFN only includes spatial structure indirectly, through the available features, we expected the performance of OK to be correlated more strongly with $MI_y$ than TabPFN. Additionally, there was even a slight negative association with the $\Delta_{\text{TabK-Tab}}R^2$ (Fig. 6b), meaning that KpR was more useful in datasets with a low $MI_y$ than in cases with a high $MI_y$. While we expected the opposite, we avoid further interpretations, as the correlation of $\Delta_{\text{TabK-Tab}}R^2$ with $MI_y$ was rather weak and non-significant ($p = 0.17$).

In contrast, neither the variance of the local MI ($\text{var}(MI_y)$) nor the number of PSSs was strongly correlated with the absolute $R^2$ (Fig. 6c, e). Yet, these two dataset properties were moderately but significantly negatively correlated with the $\Delta_{\text{TabK-Tab}}R^2$ (Fig. 6d, f). In other words, for prediction tasks with low $\text{var}(MI_y)$ or a limited number of PSSs, using KpR features significantly improved predictions.

A correlation with the number of PSSs was to be expected and has important implications for PA (Section 3.3). When features were derived from only a single PSS, the information of the feature space tended to be narrowly focused on a few characteristics (Pozzuto et al., 2025). This limited information from a single PSS will usually not suffice for accurate prediction of various different target soil properties. In these cases, integrating spatial context through KpR enabled the model to fall back on the spatial structure captured by the available soil observations, thereby improving predictive performance. However, with sensor fusion (i.e., by using more PSSs) the chance was increased that at least some of the available features were meaningfully related to the target soil properties (Ji et al., 2019; Schmidinger et al., 2024a). In such cases, the information provided by KpR features tended to be often redundant. But even when the number of PSSs was high, KpR features occasionally enhanced the model accuracy (Fig. 6f).

A low $\text{var}(MI_y)$ indicates that the spatial autocorrelation was constant throughout the study area, which aligns with the underlying second-order stationarity assumption central to OK. Hence, KpR features tended to be more valuable if the estimated spatial autocorrelation adhered to this assumption. Conversely, a high $\text{var}(MI_y)$ likely violated this assumption, thereby undermining the usefulness of KpR features.

### 3.2.2. Model characteristics

Fig. 7 illustrates the relationship between selected model characteristics with $\Delta_{\text{TabK-Tab}}R^2$. $\Delta_{\text{TabK-Tab}}R^2$ was most strongly correlated with $\Delta_{\text{Tab-OK}}R^2$, which is the difference between $R^2_{\text{TabPFN}}$ and $R^2_{\text{OK}}$, as well as $MI_{y-\hat{y}_{\text{TabPFN}}}$, which represents the spatial autocorrelation of TabPFN's test fold errors. Additionally, a modest correlation was observed with the absolute value of $R^2_{\text{TabPFN}}$. No relations were found with two further model characteristics (see Appendix B, Fig. B.2).

The strong negative correlation of $\Delta_{\text{TabK-Tab}}R^2$ with $\Delta_{\text{Tab-OK}}R^2$ was also expected (Fig. 7a). It means that TabPFN-KpR was most useful when TabPFN was already outperformed by OK. Because KpR features were derived from OK, KpR ultimately encoded more useful information in these cases than the common sensing features used in TabPFN, leading to an increased robustness. This is demonstrated by the fact that TabPFN-KpR outperformed OK in 17 out of 18 prediction tasks, while the TabPFN baseline was better than OK in 14 out of 18 prediction tasks (Fig. 5). Nonetheless, there were also three cases, in which TabPFN-KpR provided considerable improvements even when OK was on a par or slightly worse than TabPFN (Fig. 7a). Hence, TabPFN-KpR did not just mirror the performance of OK. Instead, KpR features often had a positive synergy with sensing features, as both provided unique and complementary information.

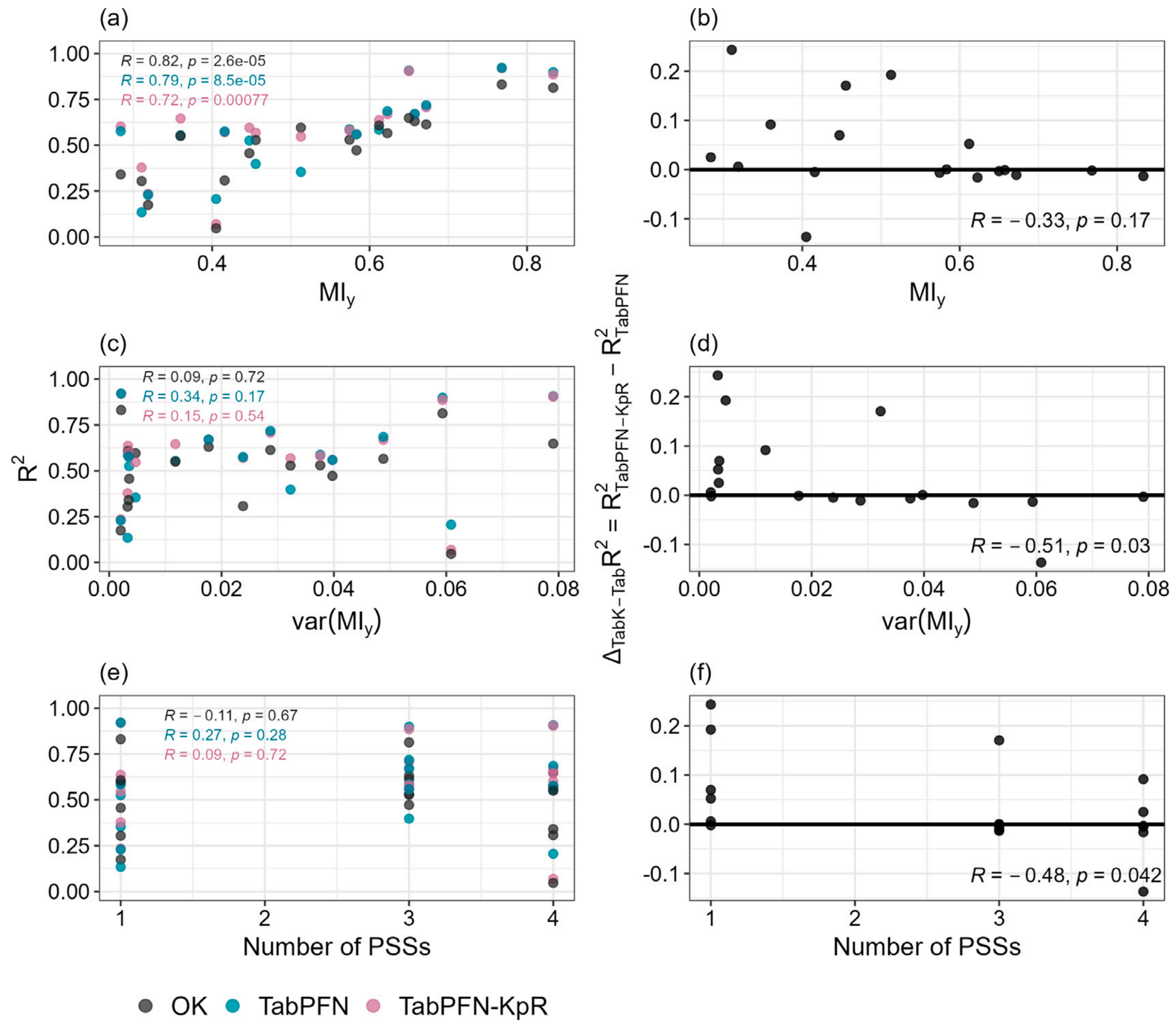

**Fig. 6.** Selected dataset characteristics and their relation to the $R^2$ of OK, TabPFN, TabPFN-KpR (a, c, e) and their relation to the $\Delta_{\text{TabK-Tab}}R^2$ (b, d, f).

The strong correlation between $\Delta_{\text{TabK-Tab}}R^2$ and the spatial autocorrelation of the TabPFN test fold errors (i.e., $MI_{y-\hat{y}_{\text{TabPFN}}}$) is fundamental for spatial ML (Fig. 7b). A high $MI_{y-\hat{y}_{\text{TabPFN}}}$ means that the sensing features could not capture certain remaining spatial dependencies (Jemeļjanova et al., 2024; Milà et al., 2024). Using spatially-explicit features (e.g., KpR) prevents this so that TabPFN-KpR proved to be particularly useful when $MI_{y-\hat{y}_{\text{TabPFN}}}$ was high. In contrast, little to no improvement was found with TabPFN-KpR, if the $MI_{y-\hat{y}_{\text{TabPFN}}}$ was low. This strong correlation of $MI_{y-\hat{y}_{\text{TabPFN}}}$ with $\Delta_{\text{TabK-Tab}}R^2$ was apparent despite the limitation associated with the $MI_{y-\hat{y}_{\text{TabPFN}}}$ as analytical tool for measuring the test-error autocorrelation (Nowosad and Meyer, 2025).

When $R^2{}_{\text{TabPFN}}$ was low, using KpR was more likely to be beneficial (Fig. 7c). Whereas when sensing features already explained most of the variability, $\Delta_{\text{TabK-Tab}}R^2$ tended to be slightly below 0. Overall, this was only a slight tendency, and as previously mentioned, most determining was the $\Delta_{\text{Tab-OK}}R^2$. For instance, in the case of RP.62 for pH predictions, $R^2{}_{\text{TabPFN}}$ was low but $R^2{}_{\text{Ok}}$ even lower, which led to overfitting and a decrease in model performance.

### 3.3. Implications for precision agriculture

Much research in PA has been dedicated to data fusion with various types of PSSs (Vullaganti et al., 2025). While this can improve the robustness of multivariate target predictions, it also increases costs. Given that the adoption of DSM in agriculture is limited by its high initial price (Lowenberg-DeBoer, 2019), data fusion with multiple PSSs is usually not within the budget of farmers. Using common sensing features alongside KpR features is also a type of data fusion but it comes with no direct extra cost. Ultimately, KpR may not be as effective as PSS fusion when comparing the total increase in accuracy but it has the potential to significantly improve the robustness of ML in PA, if sensing from only a single PSS is affordable.

We only evaluated soil properties that can be relatively well predicted with common PSSs (SOC, clay and pH). This is usually less often the case for several relevant soil nutrients (Wenjun et al., 2014; Ji et al., 2019). Hence, it is likely that KpR might possibly be more beneficial for soil nutrient mapping but this needs to be further investigated.

One drawback of KpR could be its reliance on sample sizes to reliably estimate the semivariogram. In many practical PA applications, even fewer than 50 training samples are available for the model training (Žížala et al., 2024; Vogel et al., 2025). While the sample size was not a relevant factor in this study (Appendix B; Fig. B.1f), fewer than 50 samples are usually deemed insufficient for estimating a reliable semivariogram (Webster and Oliver, 1992). In such situations, simpler spatial methods could be preferred (Karp et al., 2024) but this was outside the scope of this study. On the other hand, outside a PA context, very large sample sizes can make KpR computationally burdensome, such that nested k-fold strategies may be preferable to the resource-intensive LOO scheme. Further, the semivariogram depends on the sample configuration. For example, the regular grid sampling used in this study does not allow for reliable estimation of the nugget (Oliver and Webster, 2015). Lastly, KpR features cannot be used for extrapolation (Milà et al., 2024), which limits the applicability of KpR for field-independent predictions (Vogel et al., 2025).

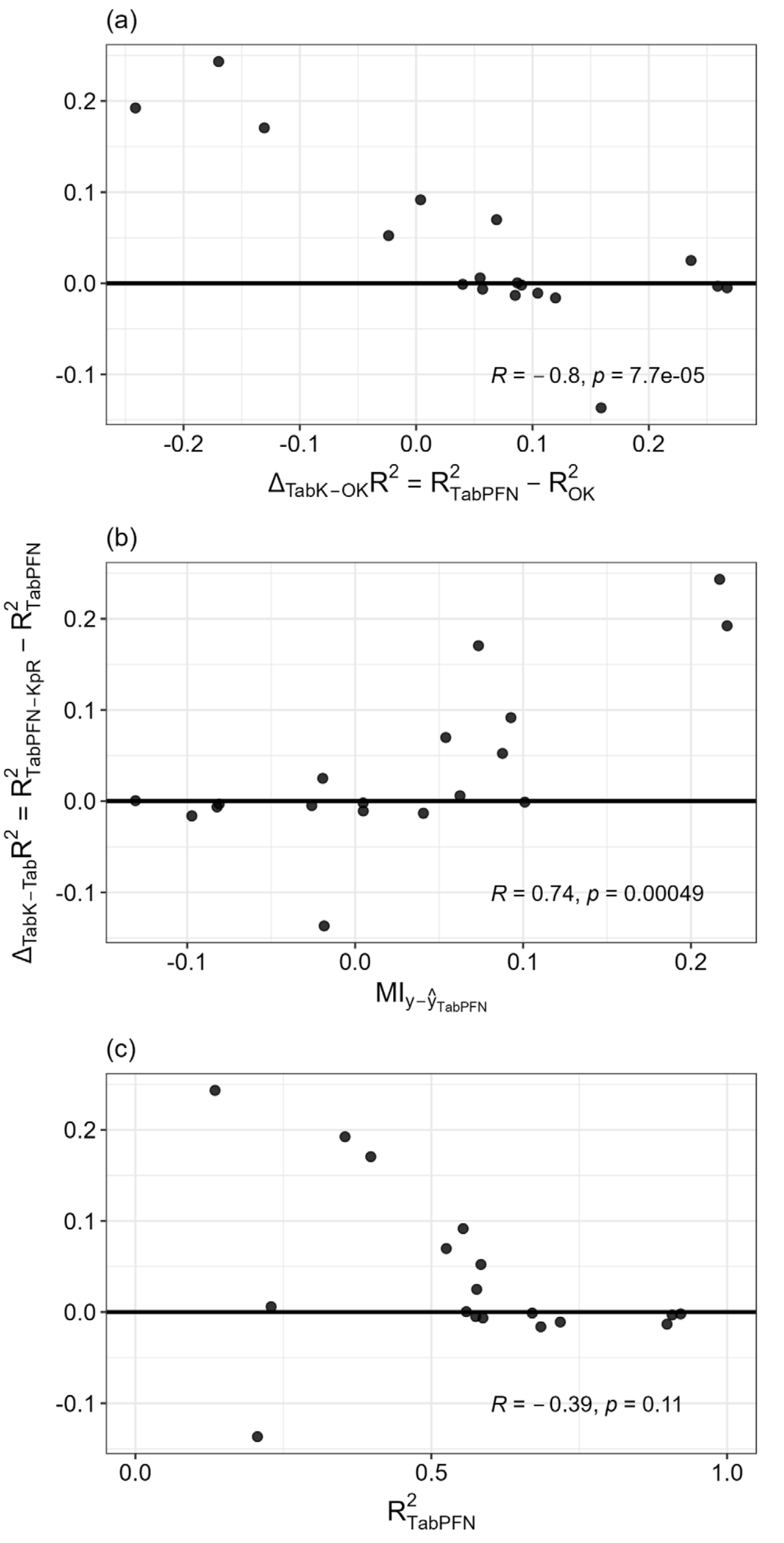

**Fig. 7.** Selected model characteristics and their relation to the $\Delta_{\text{TabK−Tab}} R^2$.

### 3.4. Further considerations

KpR is a straightforward and model-agnostic approach. This flexibility allows it to be readily applied to a wide range of models, including TabPFN. However, it may be considered relatively ad hoc (Saha et al., 2021; Heuvelink and Webster, 2022), since the spatial dependency is not inherently considered within the ML architecture or its parameter estimation. In comparison, generalized least squares may provide a more rigorous statistical framework (Saha et al., 2021; Milà et al., 2024; Zhan and Datta, 2024), but its application is bound to the model architecture and thus neither straightforward nor available in the foreseeable future for emerging models like TabPFN. Given the strong empirical performance of KpR and its solid theoretical grounding compared to previously used IpR strategies, we consider it a successful strategy for adding spatial context to ML.

We did not conduct an in-depth analysis of why TabPFN-KpR (or IpR more generally) provided more accurate point predictions than TabPFN-RK. As shown in Fig. 7b, TabPFN-KpR, like TabPFN-RK, largely accounted for the spatial autocorrelation of the residuals. While speculative, this difference may stem from interactions between sensor and kriging features, whereas RK relies on a fixed stationarity assumption for the residuals. At this stage, the observed superiority remains empirical.

An alternative approach for modelling spatial dependencies may be given by graph neural networks (Interdonato et al., 2019). Here, recent advances have also been made for spatial datasets (Bloemheuvel et al., 2023). However, in this study we focused on tabular learning, motivated by the recent promising results of TabPFN (Barkov et al., 2025).

## 4. Conclusions

Across various field-scale datasets from a precision liming context, we explored how KpR features combined with TabPFN can improve the prediction accuracy compared to other commonly used frameworks and algorithms. Our conclusions are:

- TabPFN with KpR features provided overall the most accurate predictions. We attribute this to the strong performance of the TabPFN algorithm itself for small datasets and complementary spatial context through KpR. Compared to four other commonly used ML algorithms without explicit spatial context, TabPFN-KpR meaningfully improved the mean $R^2$ by ≈30%.
- The TabPFN algorithm provided reliable uncertainty estimates for TabPFN-KpR. In contrast, the uncertainty estimates from the hybrid geostatistical framework of TabPFN-RK proved to be systematically overoptimistic. This could have unaccountable adverse effects if used in agronomic decision-making.
- KpR tended to be most beneficial for datasets with features from only a single PSS or when the spatial autocorrelation of the target soil property was stationary.
- When the test errors of a ML model are autocorrelated, or OK provides better estimates than ML, KpR features can generally be expected to boost the prediction accuracy.

## CRediT authorship contribution statement

**Jonas Schmidinger:** Writing – review & editing, Writing – original draft, Visualization, Methodology, Formal analysis, Data curation, Conceptualization. **Viacheslav Barkov:** Writing – review & editing, Software. **Sebastian Vogel:** Writing – review & editing, Supervision. **Martin Atzmueller:** Writing – review & editing, Supervision. **Gerard B.M. Heuvelink:** Writing – review & editing, Methodology, Conceptualization.

## Data and code availability

The supporting code can be found on GitHub at https://github.com/JonasSchmidinger/Kriging_prior_Regression. Supporting data is accessible through the LimeSoDa R and Python packages or Zenodo https://zenodo.org/records/14936177. However, we do not have the rights to share the coordinates of RP.62 and SA.112.

## Funding

This research was supported by the Lower Saxony Ministry of Science and Culture (MWK), funded through the zukunft.niedersachsen program of the Volkswagen Foundation, Germany (ZN4072).

## Declaration of competing interest

The authors declare that they have no known competing financial interests or personal relationships that could have appeared to influence the work reported in this paper.

## Appendix A. Benchmarking on spatial techniques

### A.1. Benchmarking materials and methods

For the extended benchmarking we compared eight additional spatial techniques in combination with TabPFN. Technical descriptions of these methods are kept to a minimum but further details are available in the cited references or our GitHub repository.

Using coordinates as features (TabPFN-Coordinates) is a simple yet established approach in spatial ML (Jemeļjanova et al., 2024). Behrens et al. (2018) expanded on that same principle by using the distance to the corners and the centre of the dataset as features. They refer to this as Euclidean distance fields (EDF). Since our study areas lack strictly defined boundaries, we defined the corners for our TabPFN-EDF implementation based on the minimum and maximum coordinate values of the available soil observations.

A model stacking framework was tested, in which we combined OK and TabPFN (Stack:TabPFN+OK) using a linear meta-model (Stack: TabPFN+OK). Predictions from OK and TabPFN were obtained via a nested three-fold CV. While a spatially weighted stacking approach, as proposed by Erdogan Erten et al. (2022), may be more appropriate in principle, its implementation was not feasible in our case due to the unavailability of code. Moreover, the use of grid-sampling based datasets reduced the necessity for spatial weighting (Fig. 2).

Sekulić et al. (2020) introduced a framework in which the feature space is augmented with a KNN distance and observation matrix. While originally proposed specifically for random forest as ‘random forest spatial interpolation’, we adapted this approach for use with TabPFN and refer to it as ‘TabPFN-KNN Matrix’. The ideal number of neighbours was tuned in a nested three-fold CV, where we tested 2 to 10 neighbours.

Liu et al. (2022) introduced the KNNpR approach to the soil domain, which involves computing the simple average of $k$-nearest neighbours to generate a single feature. Chen et al. (2024) additionally used IDW for its weights. They specifically do not refer to this technique as IDW but ’soil spatial neighbour information’. For consistency with the other methods, we refer to it as IDWpR. The same tuning strategy for the ideal number of neighbours was used as for TabPFN-KNN Matrix.

We tested LOO CV for RK scheme for (TabPFN-LOO RK). Unlike in standard RK, where residuals are computed on the training set, we used the prediction errors from a LOO CV to obtain independent inner-test errors. This approach prevents data leakage, as the inner test errors are less susceptible to overfitting compared to training residuals. To the best of our knowledge, this specific adaptation of RK has not been previously proposed.

Lastly, we used TabPFN-KpR without $\sigma^2$ (TabPFN-KpR$_{-\sigma^2}$). This means we reduced the proposed KpR framework in Eq. (6) by extending $X$ only with $\hat{y}_{\mathrm{OK}}$ but omitting $\sigma^2_{\mathrm{OK}}$.

**Table A.1**
Summary statistics of the $R^2$ over all prediction tasks for all spatial techniques combined with TabPFN.

| Model | Mean $R^2$ | Median $R^2$ | SD $R^2$ |
|---|---|---|---|
| TabPFN-KNN Matrix | 0.578 | 0.569 | 0.224 |
| Stack:TabPFN+OK | 0.584 | **0.601** | 0.233 |
| TabPFN-RK | 0.585 | 0.582 | 0.209 |
| TabPFN-LOO RK | 0.585 | 0.590 | 0.217 |
| TabPFN-Coordinates | 0.586 | 0.592 | 0.210 |
| TabPFN-EDF | 0.593 | 0.587 | 0.199 |
| TabPFN-KNNpR | 0.596 | 0.587 | 0.205 |
| TabPFN-KpR | 0.597 | 0.598 | 0.213 |
| TabPFN-IDWpR | 0.599 | 0.587 | 0.204 |
| TabPFN-KpR$_{-\sigma^2}$ | **0.600** | 0.595 | 0.204 |

### A.2. Benchmarking results and discussions

As evident from Table A.1, all spatial techniques improved the $R^2$ compared to the non-spatial TabPFN baseline (Table 2). Hence, incorporating spatial context proved generally beneficial, through different strategies. The only exception was TabPFN-KNN Matrix, the lowest-performing spatial method, which returned a slightly lower median $R^2$.

The IpR techniques, i.e., TabPFN-KpR, TabPFN-KpR$_{-\sigma^2}$, TabPFN-IDWpR and TabPFN-KNNpR led on average to greater accuracy improvements than the other techniques. Stack:TabPFN+OK had a higher median $R^2$ but a relatively low mean $R^2$ in comparison. The overall performance difference between the IpR techniques was minor. TabPFN-IDWpR and TabPFN-KpR$_{-\sigma^2}$ had a marginally higher mean $R^2$ than TabPFN-KpR but had a lower median $R^2$.

Given the theoretical superiority of OK over IDW (Goovaerts, 2000), we expected KpR to outperform IDWpR. While this did not materialize, the number of datasets may have been too small to detect significant differences, especially since both methods operate on similar principles within the IpR framework. Nonetheless, beyond its stronger theoretical foundation, KpR offers practical advantages: it does not require tuning the number of neighbours, making it more computationally efficient and straightforward to apply. Lastly, KpR incorporates adding $\sigma^2_{\mathrm{OK}}$ as an additional feature. However, this component was insignificant for our datasets based on grid-sampling (Fig. 2), where $\sigma^2_{\mathrm{OK}}$ would be nearly constant. For this reason, TabPFN-KpR did not outperform TabPFN-KpR$_{-\sigma^2}$. In the case of spatially clustered datasets and thus non-constant $\sigma^2_{\mathrm{OK}}$ values, it would likely become a more influential factor. Exploring this further was outside the scope of this study, as we did not have access to enough datasets with spatially clustered samples. Further, the validation would have been less straightforward (Wadoux et al., 2021).

TabPFN-LOO RK did not result in significantly more accurate predictions than TabPFN-RK. However, it provided improved uncertainty estimates (Appendix B.3, Fig. B.3).

## Appendix B. Extended results and discussions

### B.1. Performance per dataset

Table B.1 shows the performance of the four main models for each individual prediction task. As discussed in the paper, there was one outlier for TabPFN-KpR in the case of pH in RP.62. Compared to the TabPFN baseline, TabPFN-KpR was significantly worse in this prediction task. On the other hand, TabPFN-KpR improved the $R^2$ of SOC in SA.112 by 0.10 compared to both OK and TabPFN, hinting at complementary information from the geostatistical features.

**Table B.1**
Performance of the main four models for each prediction task.

| Dataset | Soil property | OK | | | TabPFN | | | TabPFN-RK | | | TabPFN-KpR | | |
|---|---|---|---|---|---|---|---|---|---|---|---|---|---|
| | | $R^2$ | RMSE | ME | $R^2$ | RMSE | ME | $R^2$ | RMSE | ME | $R^2$ | RMSE | ME |
| BB.250 | SOC | 0.813 | 0.212 | −0.005 | 0.898 | 0.156 | −0.006 | 0.904 | 0.152 | −0.001 | 0.885 | 0.166 | −0.013 |
| | Clay | 0.613 | 1.723 | −0.005 | 0.718 | 1.471 | 0.014 | 0.720 | 1.466 | 0.002 | 0.707 | 1.499 | −0.122 |
| | pH | 0.631 | 0.305 | 0.000 | 0.671 | 0.288 | 0.002 | 0.679 | 0.284 | 0.001 | 0.670 | 0.288 | 0.003 |
| G.150 | SOC | 0.596 | 0.064 | 0.000 | 0.354 | 0.081 | 0.001 | 0.510 | 0.070 | 0.000 | 0.547 | 0.067 | −0.003 |
| | Clay | 0.174 | 3.049 | −0.016 | 0.229 | 2.946 | 0.034 | 0.209 | 2.984 | 0.002 | 0.235 | 2.935 | −0.010 |
| | pH | 0.456 | 0.220 | −0.001 | 0.525 | 0.205 | −0.007 | 0.544 | 0.201 | −0.003 | 0.595 | 0.190 | −0.003 |
| MG.112 | SOC | 0.608 | 0.162 | 0.002 | 0.584 | 0.167 | 0.002 | 0.634 | 0.156 | −0.005 | 0.636 | 0.156 | −0.010 |
| | Clay | 0.831 | 4.875 | 0.176 | 0.922 | 3.320 | −0.086 | 0.922 | 3.310 | −0.040 | 0.920 | 3.363 | 0.121 |
| | pH | 0.304 | 0.230 | −0.016 | 0.135 | 0.257 | 0.003 | 0.381 | 0.217 | −0.010 | 0.378 | 0.218 | 0.002 |
| SA.112 | SOC | 0.550 | 0.165 | 0.002 | 0.554 | 0.165 | 0.005 | 0.587 | 0.158 | 0.000 | 0.645 | 0.147 | −0.011 |
| | Clay | 0.308 | 1.642 | 0.053 | 0.575 | 1.286 | −0.082 | 0.544 | 1.333 | −0.008 | 0.570 | 1.294 | 0.019 |
| | pH | 0.340 | 0.323 | 0.008 | 0.577 | 0.259 | −0.004 | 0.578 | 0.259 | −0.002 | 0.602 | 0.251 | −0.005 |
| BB.72 | SOC | 0.472 | 0.240 | −0.010 | 0.559 | 0.219 | −0.020 | 0.524 | 0.228 | −0.003 | 0.560 | 0.219 | −0.030 |
| | Clay | 0.530 | 1.952 | −0.040 | 0.587 | 1.830 | 0.054 | 0.597 | 1.809 | 0.034 | 0.581 | 1.844 | 0.059 |
| | pH | 0.528 | 0.387 | 0.003 | 0.398 | 0.438 | −0.019 | 0.412 | 0.432 | −0.017 | 0.568 | 0.370 | 0.004 |
| RP.62 | SOC | 0.565 | 0.268 | −0.010 | 0.685 | 0.228 | −0.000 | 0.688 | 0.227 | 0.005 | 0.669 | 0.234 | −0.011 |
| | Clay | 0.648 | 4.972 | −0.122 | 0.907 | 2.553 | 0.154 | 0.908 | 2.541 | 0.140 | 0.904 | 2.597 | −0.014 |
| | pH | 0.047 | 0.224 | 0.016 | 0.206 | 0.205 | 0.005 | 0.187 | 0.207 | 0.011 | 0.069 | 0.221 | 0.015 |

### B.2. All dataset and model characteristics

In the main analysis, we focused only on those dataset and model properties that were at least slightly correlated with $\Delta_{\text{TabK-Tab}}\text{R}^2$. In this section, we present all further characteristics that showed no or minimal correlations with $\Delta_{\text{TabK-Tab}}\text{R}^2$.

Fig. B.1 shows four further dataset characteristics. Apart from the skewness, all these properties were slightly to strongly correlated with the $\text{R}^2$ but none of these properties had a meaningful correlation with the $\Delta_{\text{TabK-Tab}}\text{R}^2$. Fig. B.2 shows the $\Delta_{\text{TabK-Tab}}\text{R}^2$ from two remaining model characteristics, without any notable pattern.

### B.3. Overoptimistic uncertainty estimates in regression kriging

Fig. B.3 shows the QCP reliability plot of TabPFN-LOO RK and TabPFN-RK. As evident from that plot, the uncertainty estimates of TabPFN-LOO RK are more reliable and not systematically overoptimistic. We attribute this to the fact that in TabPFN-RK, $\sigma^2_{\text{RK}}$ is estimated from the training residuals. $\sigma^2_{\text{RK}}$ decreases as the degree of overfitting on the training data increases. E.g., in the case of total overfitting with zero residuals on the training data, $\sigma^2_{\text{RK}}$ would boil down to zero:

$$e_{\text{ML}}(s_1) = \cdots = e_{\text{ML}}(s_n) = 0 \implies \sigma^2_{\text{RK}} = 0. \quad \text{(B.1)}$$

Hence, we argue that RK will often have the tendency to be overoptimistic, as it is vulnerable to the degree of overfitting on the training dataset.

In the case of TabPFN-LOO RK, this problem was avoided. Even if the training residuals are zero due to overfitting, the inner test errors from the nested CV are interpolated to adjust the final predictions. The inner test errors would capture the degree of overfitting, so that $\sigma^2_{\text{LOO RK}} \neq 0$, apart from the unrealistic case where the residuals are zero due to a perfect model.

### B.4. Semivariogram analysis

The semivariogram models for all datasets and soil properties are presented in Fig. B.4. Note that these semivariograms were computed and fitted using all the data, while in the actual analysis we computed semivariograms always on only 90% of the data, due to the 10-fold cross-validation approach used. In some instances, the fitted semivariogram model does not align that well with the empirical semivariogram (for example, for Clay and SOC in MG.112) when assessed "by eye". However, *automap* selects from a predefined set of semivariogram model functions and estimates the model parameters by minimizing a loss function. We employed the default *automap* and *gstat* settings, considering spherical, exponential, Gaussian and Matérn models, and a minimization loss function based on distance and number of sample pairs.

## Appendix C. Notes on data leakage

The authors of Ohmer et al. (2025) raised concerns about spatial lag features, which we refer to as IpR, due to problems with data leakage. However, we argue that this is a misconception and IpR is a valid approach, apart from the limitations listed in Milà et al. (2024).

The concern of data leakage may arise from the fact that we use the target soil property to predict itself. This could appear as circular logic. However, this 'circularity' is the base of every spatial interpolation technique Section 2.3.1, which predicts values at unsampled locations given neighbouring observations. IpR features operate on the same principle with a slightly different end-product. In a spatial interpolation, the final prediction is the product; in IpR, the interpolation prediction becomes one of several inputs to the ML model. Therefore, we are not using an observation's target to predict itself but information from surrounding spatial observations as additional spatial context. This is relatively analogous to model stacking, where the output of one model becomes input for another.

Problems with data leakage may occur if the IpR features are constructed prior to the test and train separation. In this case, we contaminate the features with direct knowledge from the test samples, which would greatly inflate the model performance and deteriorate the model generalization. However, such a mistake would generally be a textbook example of data leakage regardless of IpR (Kapoor and Narayanan, 2023).

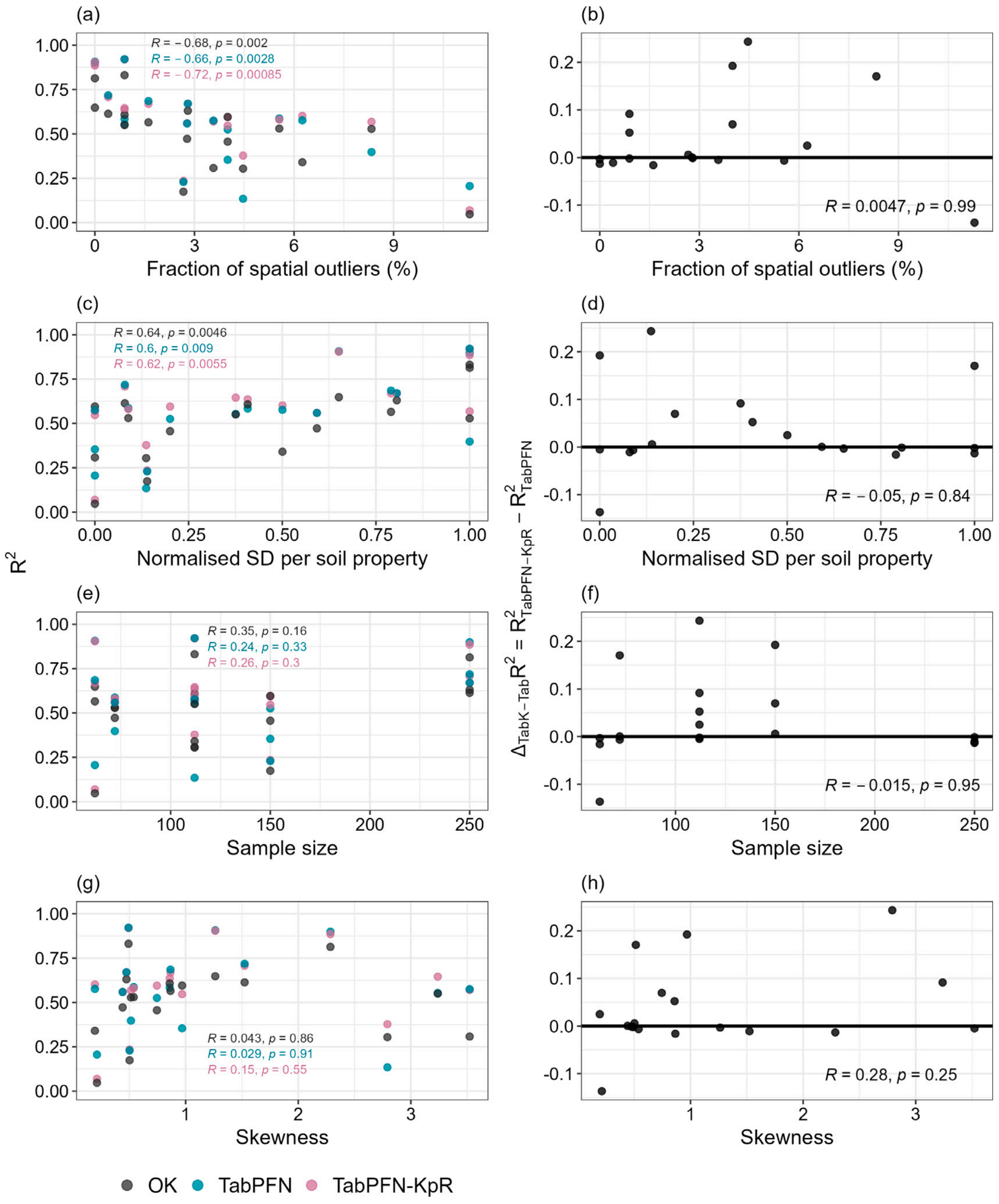

**Fig. B.1.** Remaining dataset characteristics and their relation to the $R^2$ of OK, TabPFN, TabPFN-KpR (a,c,e,g) and their relation to the $\Delta_{TabK-Tab}R^2$ (b,d,f,h).

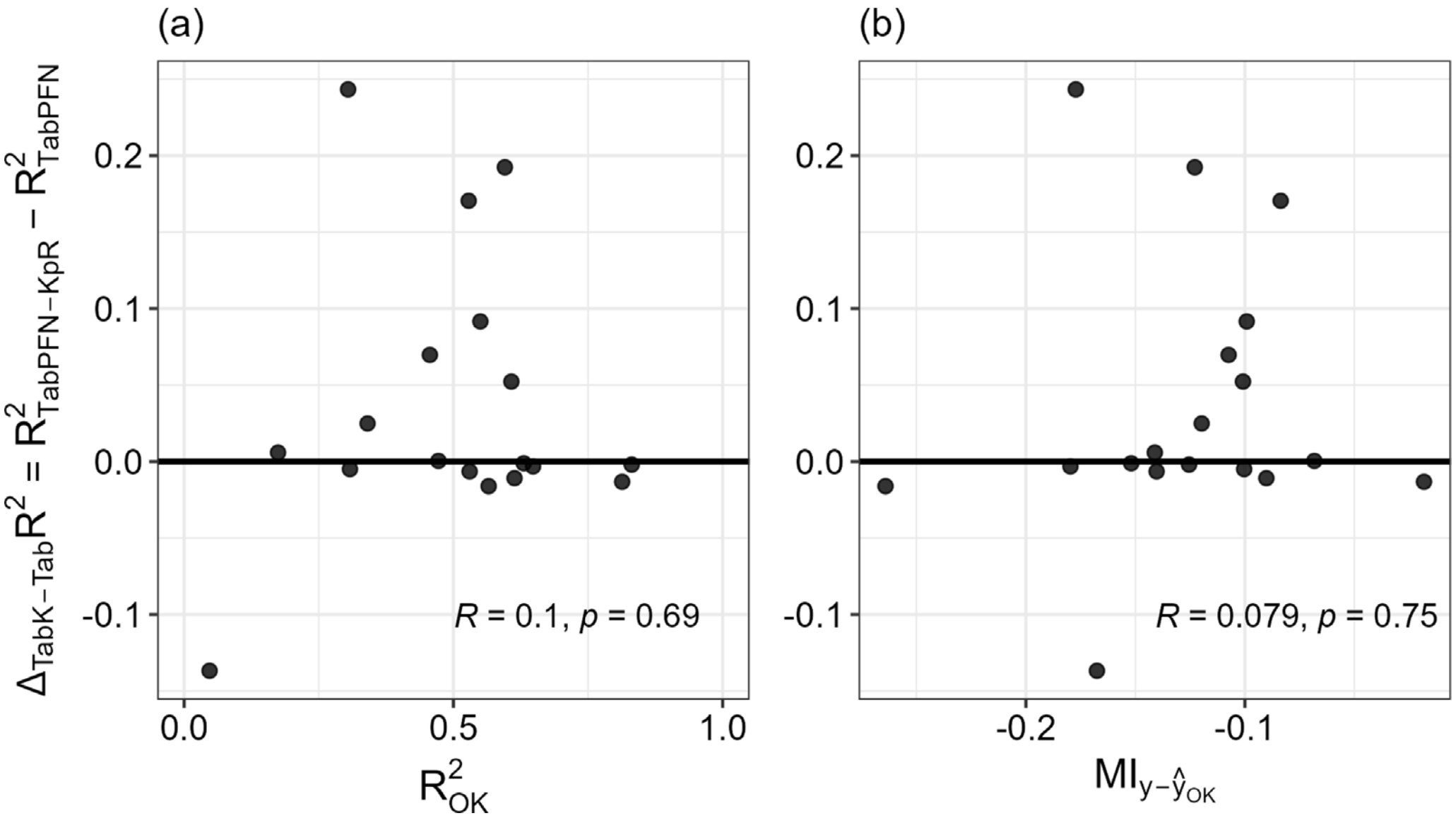

**Fig. B.2.** Remaining model characteristics and their relation to the $\Delta_{\mathrm{TabK-Tab}}\mathrm{R}^2$.

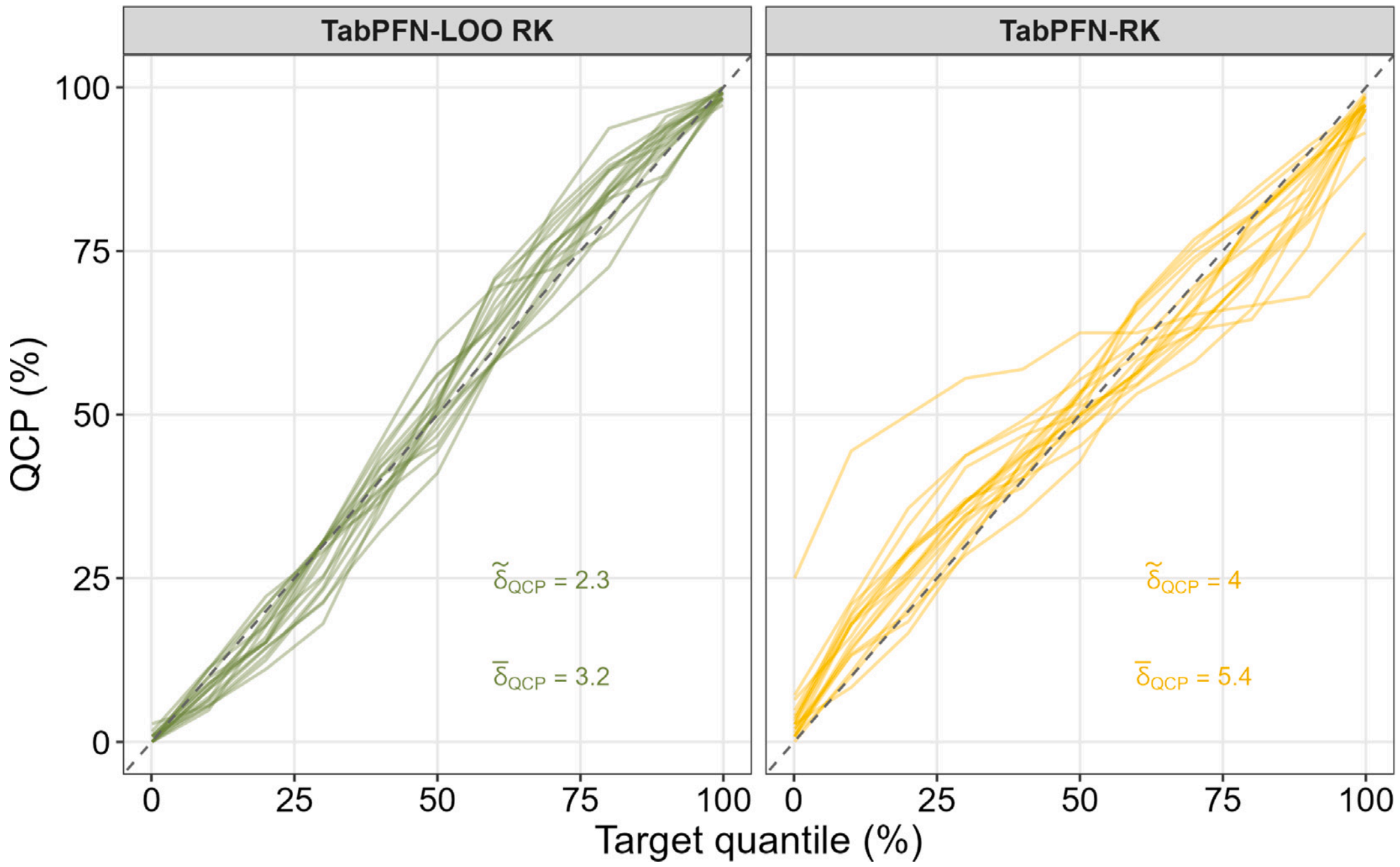

**Fig. B.3.** Reliability plots of the quantile coverage with TabPFN-LOO RK and TabPFN-RK. Each line represents the QCP obtained from one of 18 prediction tasks. Lines closer to the 1:1 line indicate more reliable uncertainty estimates. The mean and median deviation from the 1:1 line is measured by the $\bar{\delta}_{\mathrm{QCP}}$ and $\tilde{\delta}_{\mathrm{QCP}}$.

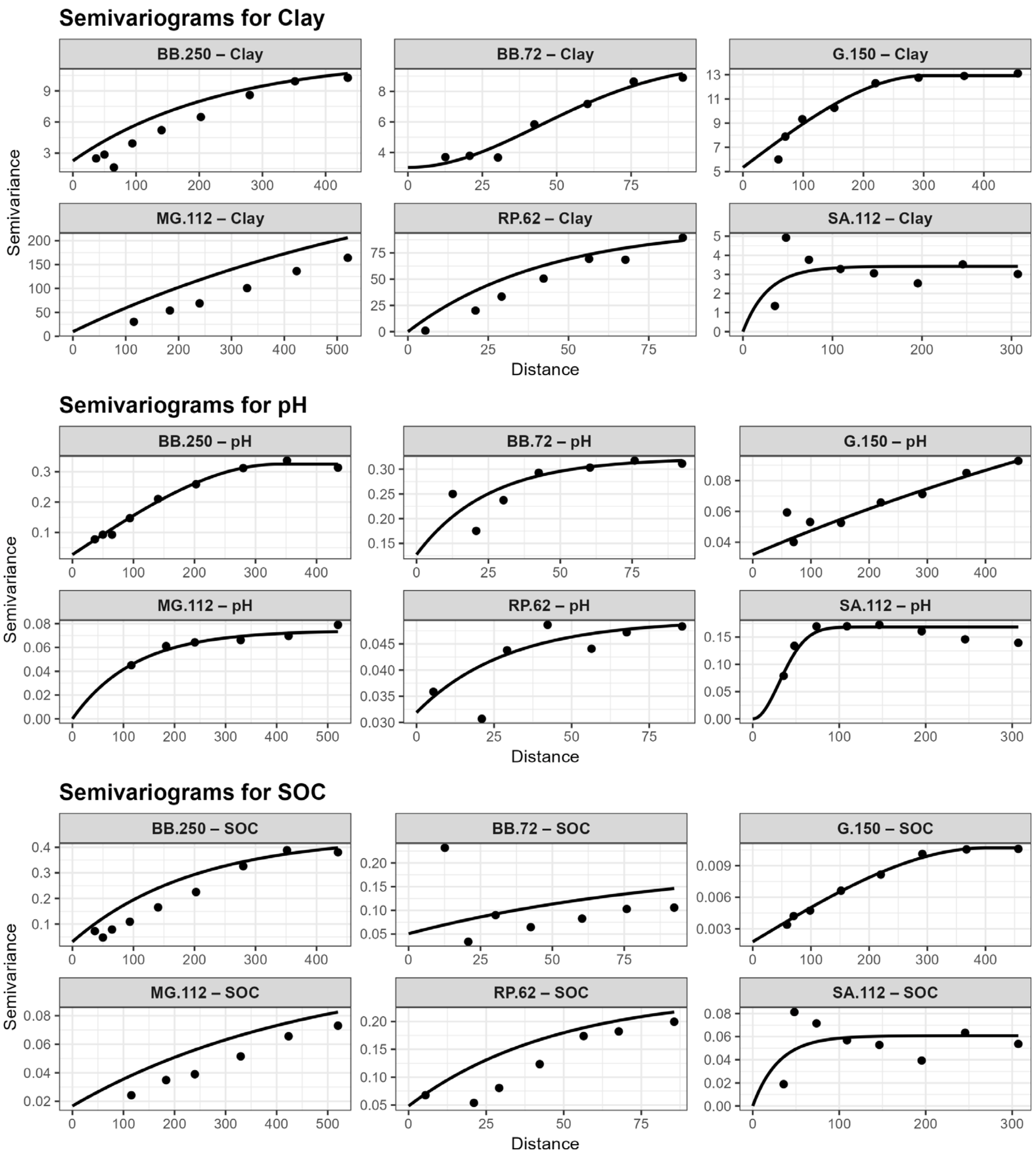

**Fig. B.4.** Semivariogram models of every dataset and soil property based on the whole sample set.

## Data availability

The supporting code can be found on GitHub at https://github.com/JonasSchmidinger/Kriging_prior_Regression. Supporting data is accessible through the LimeSoDa R and Python packages or Zenodo https://zenodo.org/records/14936177. However, we do not have the rights to share the coordinates of RP.62 and SA.112.